\documentclass[journal]{IEEEtai}

\usepackage[colorlinks,urlcolor=blue,linkcolor=blue,citecolor=blue]{hyperref}

\usepackage{color,array}

\usepackage{graphicx}

\usepackage{soul}
\usepackage{multirow}
\usepackage{graphicx}
\usepackage{subfig}
\usepackage{todonotes}
\usepackage{xcolor,colortbl}
\usepackage{arydshln}

\let\labelindent\relax
\usepackage{enumitem}

\usepackage{tikz}
\usepackage{pgfplots}
\usepgfplotslibrary{groupplots}
\usepackage{amsmath}
\usepackage{amssymb}
\usepackage{times}
\usepackage{latexsym}
\usepackage[T1]{fontenc}
\usepackage[utf8]{inputenc}
\usepackage{microtype}
\usepackage{wrapfig}
\usepackage{booktabs, multirow}
\usetikzlibrary{shapes,snakes,positioning,backgrounds,calc,tikzmark, fit}

\definecolor{LightCyan}{rgb}{0.88,1,1}

\newcommand{\name}{\textsf{TGIF}}
\newcommand{\citet}{\cite}

\begin{document}

\title{Emotion Flip Reasoning in Multiparty Conversations}

\author{Shivani Kumar, Shubham Dudeja, Md Shad Akhtar, Tanmoy Chakraborty
\thanks{28th November 2022. This work was supported by ihub-Anubhuti-iiitd Foundation, set up under the NM-ICPS scheme of the DST.}
\thanks{Shivani Kumar is with Indraprastha Institute of Information Technology Delhi, India. (e-mail: shivaniku@iiitd.ac.in).}
\thanks{Shubham Dudeja was with Indraprastha Institute of Information Technology Delhi, India. (e-mail: shubham19053@iiitd.ac.in).}
\thanks{Md Shad Akhtar is with Indraprastha Institute of Information Technology Delhi, India. (e-mail: shad.akhtar@iiitd.ac.in)}
\thanks{Tanmoy Chakraborty is with Indian Institute of Technology Delhi, India. (e-mail: tanchak@iitd.ac.in)}}

\markboth{Journal of IEEE Transactions on Artificial Intelligence, Vol. 00, No. 0, Month 2023}
{Kumar \MakeLowercase{\textit{et al.}}: Emotion Flip Reasoning in Multiparty Conversations}

\maketitle

\begin{abstract}
In a conversational dialogue, speakers may have different emotional states and their dynamics play an important role in understanding dialogue's emotional discourse. However, simply detecting emotions is not sufficient to entirely comprehend the speaker-specific changes in emotion that occur during a conversation. To understand the emotional dynamics of speakers in an efficient manner, it is imperative to identify the rationale or instigator behind any changes or flips in emotion expressed by the speaker. In this paper, we explore the task called Instigator based \textbf{E}motion \textbf{F}lip \textbf{R}easoning (EFR), which aims to identify the instigator behind a speaker's emotion flip within a conversation. For example, an emotion flip from \textit{joy} to \textit{anger} could be caused by an instigator like \textit{threat}. To facilitate this task, we present MELD-I, a dataset that includes ground-truth EFR instigator labels, which are in line with emotional psychology. To evaluate the dataset, we propose a novel neural architecture called \name, which leverages Transformer encoders and stacked GRUs to capture the dialogue context, speaker dynamics, and emotion sequence in a conversation. Our evaluation demonstrates state-of-the-art performance ($+4-12\%$ increase in F1-socre) against five baselines used for the task. Further, we establish the generalizability of \name\ on an unseen dataset in a zero-shot setting. Additionally, we provide a detailed analysis of the competing models, highlighting the advantages and limitations of our neural architecture.
\end{abstract}

\begin{IEEEImpStatement}
Emotions play a pivotal role in deciding the impact of a statement uttered. However, in a conversational setting, simply identifying  the emotions of  utterances in a dialogue is not enough to characterize the emotional dynamic of the speaker. To this end, the proposed task of emotion-flip reasoning is eminent. The proposed flip explanations via triggers and instigators can help scrutinise how a particular type of remark or expression affects the end listener. A response generation mechanism can use these triggers or instigators as feedback to steer a conversation so that the user feels chipper.
\end{IEEEImpStatement}


\section{Introduction}\label{sec:intro}
    Understanding emotions is essential for assessing the current state of a speaker in a conversation. Consequently, there has been a significant amount of research in this field \cite{survey}. Emotional awareness has proven beneficial in areas that involve aspect analysis of users such as social media \cite{social-media1,social-media2,social-media3}, and e-commerce \cite{e-commerce}. Initial studies focused on standalone texts like tweets \cite{social-media1,social-media2} to extract the appropriate emotions. However, with the advent of online dialogue agents, the focus of emotion analysis has shifted towards conversation data, usually termed as Emotion Recognition in Conversation (ERC) \cite{erc}. Here, the input is a sequence of utterances or a dialogue, instead of isolated texts, and the aim is to identify the emotion of each dialogue utterance. Though emotion is an imperative aspect of a conversation, we posit that it is insufficient to simply identify the speakers' emotion in a dialogue. To reason out the change/flip in  emotion of a speaker, a more detailed analysis is required. To this end, we expore the task -- {\bf Emotion Flip Reasoning (EFR)} \cite{kumar2022discovering}.

EFR deals with identifying the cause/reason behind an emotional flip of a speaker in a dialogue. The entire EFR pipeline works in three stages-
\begin{enumerate}[label=(\alph*), leftmargin=*]
    \item Given a sequence of dialogue utterances with emotion labels, the first stage of EFR identifies the utterance where a speaker experienced a flip of emotion.
    \item In the second stage, EFR identifies utterances or triggers responsible for the emotion flip.
    \item Finally, EFR assigns psychologically motivated \cite{mooren1993contributions,lazarus} instigator labels to triggers to explain the emotion flip.
\end{enumerate}
This paper focuses on the third stage, as the first stage can be effortlessly executed from a dialogue with emotion labels, while the second stage \cite{kumar2022discovering} is inherent in the third.

Following the cognitive appraisal theory \cite{lazarus}, we define a finite set of $27$ instigators to reason out flips. Here, we do not account for implicit emotion flips, i.e., emotion flips due to the absence of explicit instigator (e.g., verbal articulation). For instance, emotion flips due to reminiscence can be regarded as implicit. On the other hand, an external trigger is associated with an emotion flip that occurs due to something mentioned in the text (e.g., a person getting scolded).

The task of EFR has the capability to improve user experience in a conversational dialogue system especially in empathetic response generation \cite{dialog_agent1,dialog_agent2,dialog_agent3}. A dialogue agent can use EFR triggers for emotion flips in a feedback mechanism. It can award/penalize the agent and the response generator to repeat/avoid using a similar utterance in future conversations. Moreover, the knowledge of instigators can be used {\em to explain} such emotion flips.

\textbf{Problem Statement:}
The EFR task can be formally defined as follows:
Given a sequence of $n$ tuples of the form $\langle u_i, s_j, e_k \rangle$ in a multiparty conversation, where $s_j \in S$ is a speaker from a predefined speaker set $S$, $e_k \in E$ is a set of emotion labels, and $u_i \in D$ is an utterance of the dialogue $D$, we associate instigator label(s) with an utterance $u_l$ if it causes a flip/change in emotion of a speaker $s_m\in S$ in their consecutive utterances in the conversation. We define the objective in the following three steps.
\begin{itemize}[leftmargin=*]
    \item First, we identify all utterances $u_i$ in a dialogue which experience an emotion flip of some speaker $s_j$ -- when the emotion of the utterance 
    is different from the emotion associated with the last utterance of speaker $s_j$. We call $u_i$ as the target utterance. 
    
    \item Second, for each emotion flip utterance $u_i$, we mark a set of utterances $u_k \in \{u_1, u_2, \cdots, u_i\}$  which are responsible for the emotion flip at $u_i$. 
    
    \item Finally, we associate psychologically motivated instigator labels (c.f. Section \ref{sec:dataset}) to each $u_k$ for a target utterance $u_i$. 
\end{itemize}

\begin{figure}[t]
    \centering
    \includegraphics[width=0.85\columnwidth]{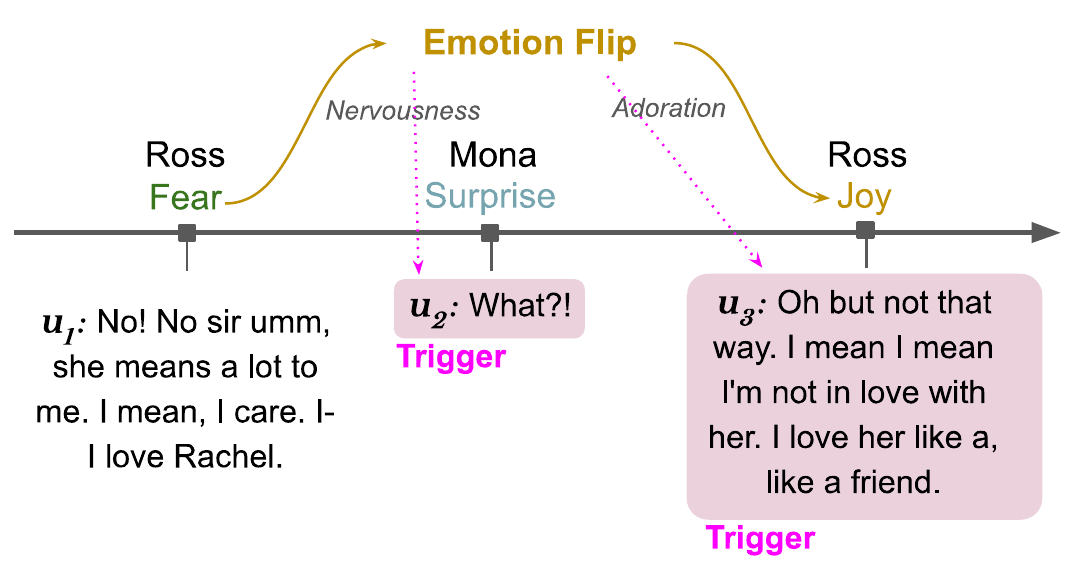}
    \caption{Example of an emotion flip with triggers and instigators. Ross's emotion flipped from \textit{Fear} ($u_1$) to \textit{Joy} ($u_3$) due to two trigger utterances ($u_2$ and $u_3$) caused by the instigators, \textit{nervousness} and \textit{adoration}, respectively.}
    \label{fig:data_example2}
    \vspace{-5mm}
\end{figure}

Figure \ref{fig:data_example2} illustrates an example of emotion flip with corresponding instigators. It shows a multiparty scenario where three speakers are engaged in a conversation. There are two emotion flips -- $\langle u_1$, $Ross$, $fear$$\rangle$ $\rightarrow$ $\langle u_3$, $Ross$, $joy$$\rangle$  and $\langle u_3$, $Ross$, $joy$$\rangle$ $\rightarrow$ $\langle u_5$, $Ross$, $anger$$\rangle$. The first flip occurs due to two trigger utterances, $u_2$  and $u_3$, each evoking the feeling of nervousness and adoration in the speaker (Ross). Consequently, the instigator labels for the concerned flip would be \textit{Nervousness} and \textit{Adoration}. On the other hand, the trigger for the second flip is a single utterance ($u_4$), and the corresponding instigator labels are \textit{Annoyance} and \textit{Challenge} as the trigger instigates the notion of annoyance and challenge in the speaker (Ross). This example highlights the case when more than one trigger utterance can cause an emotion flip.

\begin{figure}[t]
    \centering
    \includegraphics[width=\columnwidth]{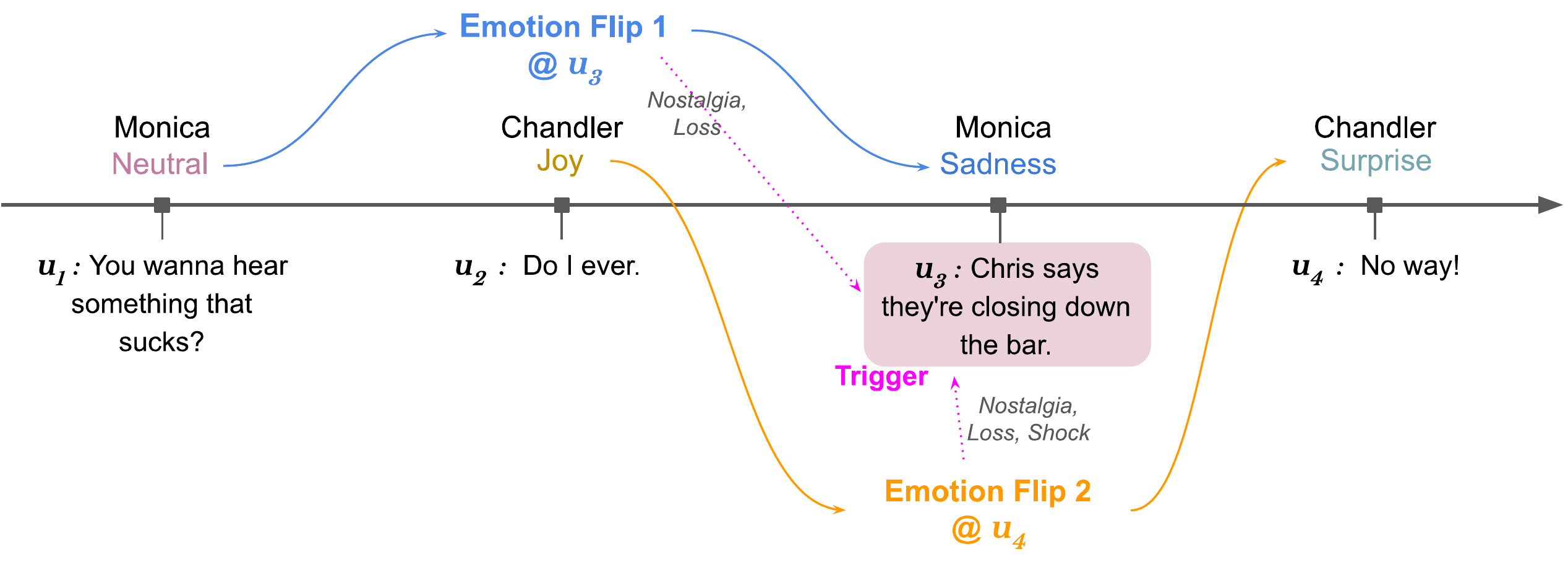}
    \caption{Example of an emotion flip with self-trigger. Monica's emotion flipped from \textit{Neutral} ($u_1$) to \textit{Sadness} ($u_3$) due to one trigger utterance ($u_3$ itself) caused by the instigators \textit{nostalgia} and \textit{loss}. The other speaker's emotion then flipped from \textit{Joy} ($u_2$) to \textit{Surprise} ($u_4$) due to a single trigger utterance ($u_3$ again) caused by the instigators \textit{nostalgia}, \textit{loss} and \textit{shock}.}
    \label{fig:data_example1}
    \vspace{-4mm}
\end{figure}

In Figure \ref{fig:data_example1}, we show another example from our dataset. It shows a dyadic conversation having two emotion flips ($u_3$ and $u_4$) corresponding to two speakers (Monica and Chandler) involved in the conversation. The emotion flip at $u_3$ is an example of a self-trigger emotion flip -- the responsible utterance (or trigger) is $u_3$ itself. Moreover, the same utterance $u_3$ acts as the trigger for both the emotion flips but causes different instigators in the two cases. It is interesting to note that the same utterance causes the emotion \textit{sadness} in one speaker while the emotion \textit{surprise} in another. This highlights the importance of identifying the emotion instigators to understand emotion dynamics completely. 

\textbf{\bf Contributions:} 
We summarise our contributions below-
\begin{itemize}[leftmargin=*]
\setlength{\itemsep}{0pt}
\setlength{\parskip}{0pt}
    \item We propose a {\bf novel task} -- Emotion Flip Reasoning (EFR) in conversational dialogue that aims to explain the shift in emotion for a speaker present in the dialogue.
    \item We carefully draft a set of ground-truth labels, called \textbf{instigators}, to explain an emotion flip.
    \item We develop a new dataset, called {\bf MELD-I}.
    \item We design \textbf{\name}, a Transformer and GRU-based architecture for the EFR task of multi-label instigator classification.
\end{itemize}

We highlight some related literature in the next section followed by the illustration of the datset used in Section \ref{sec:dataset}. Our proposed methodlogy to solve the EFR task is represented in Section \ref{sec:methodology} while the next section shows the experimental setup and results. Finally we analyse the obtained results and conclude in the last section.

The source code of {\name} and the proposed dataset, MELD-I, along with the execution instructions are available at \url{https://github.com/LCS2-IIITD/EFR-Instigators.git}.

\section{Related Work}
    \label{sec:related_work}
    {\textbf{Emotion recognition.}} Earlier studies in the field of emotion analysis \cite{ekman,picard,cowen2017self,mencattini2014speech,zhang2016intelligent,cui2020eeg} dealt with only standalone inputs. \citet{liew2016exploring} developed a dataset of $5,553$ tweets and manually annotated them with emotion labels. Further, they evaluated this dataset using standard machine learning techniques. Recently \citet{kaminska2021nearest} used the weighted k-nearest neighbour approach in order to provide an explainable model for emotion detection in tweets. Mulimodality, like speech \cite{SINGH2022245} and visual signals \cite{THUSEETHAN2022174,9516906}, is also a well explored topics in the domain. A detailed survey is provided by \cite{9736644}. However, these studies are performed for standalone text, which lacks any contextual information.

{\textbf{Emotion and conversation.}} Recently, the focus of emotion detection has shifted to conversations (emotion recognition in conversation or ERC). It has gained significant popularity due to numerous applications. Existing literature suggests that a wide range of deep learning methods have been applied to address the Emotion Recognition in Conversation (ERC) task \cite{icon,erc_knowledge,bieru,dialoguegcn,aghmn,tlerc,dialogxl,poria2017context,jiao2019higru,9706271,yang2022hybrid,ma2022multi,lian2022smin}. ICON \cite{icon} used a memory network architecture to model the interaction between self and inter-speaker states in two-party conversations. On the other hand, the use of external knowledge was explored in \citet{erc_knowledge} along with a hierarchical self-attention mechanism to detect emotions in conversation. \citet{bieru} used a party ignorant framework for conversation sentiment analysis. The use of graph convolutional networks to capture the inter-speaker dynamics in a dialogue was explored by \citet{dialoguegcn}. They utilized the dependencies among speakers to capture the contextual dynamics in an efficient way. In another work, AGHMN \cite{aghmn} proposed a hierarchical memory network with an attention mechanism to capture the essence of the dialogue in order to get a better understanding of the emotional dynamics of the speaker. TL-ERC \cite{tlerc} exploited the learned parameters of a dialogue generation module for emotion recognition through the transfer learning setup. Recently, DialogXL \cite{dialogxl} adopted XLNet \cite{xlnet} model for ERC. They encoded the dialogue utterances and made use of dialogue-aware self-attention to exploit the dialogue semantics. A hierarchical gated recurrent unit framework involving two GRUs at different levels was employed by \citet{jiao2019higru} where a lower-level GRU modeled the word-level inputs while an upper-level GRU captured the context at the utterance level. Further, \citet{LIAN2021483} proposed a correction model for previous approaches called ``Dialogical Emotion Correction Network (DECN)". The aim of this work was to improve upon the emotion recognition performance by automatically identify errors made by emotion recognition strategies. The authors proposed the use of a graphical network to model human interactions in dialogues. Another study \cite{SHOU2022629} used graph to solve the problem of ERC. They proposed a conversational affective analysis model which combined dependent syntactic analysis and graph convolutional neural networks. A self-attention mechanism captures the most effective words in the conversation, followed by graph construction. The authors shows experiments on various datasets the report higher accuracy than previous methods.

{\textbf{Beyond emotion recognition.}}
Most of the existing ERC systems do not account for the explainability of emotions. In an attempt to do so, \citet{emotion-cause1} proposed the task of emotion-cause analysis. The task deals with identifying a span in the text responsible for a specific emotion. For instance, we observe two emotions in the sentence `\textit{The queue was so long, but at last I got vaccinated}' -- \textit{joy} and \textit{disgust}. The task aims at identifying the phrases `\textit{the queue was long}' for \textit{disgust} and `\textit{I got vaccinated}' for \textit{joy}. Following this work, \citet{emotion-cause2} investigated the use of linguistic phenomenon by proposing an SVM-based model for emotion-cause identification. Xia et al. \cite{xia2019emotioncause} proposed another task- emotion-cause pair extraction. This task tried to extract the potential pairs of emotions and the corresponding causes in a document. The proposed a two-step approach where, first, individual emotion extraction and cause extraction are performed via multi-task learning and then emotion-cause pairing and filtering are done. In another one of their work \citet{xia2019rthn} proposed a joint emotion-cause extraction framework which consisted of two encoders. A RNN based encoder was employed to get the word-level representations while a Transformer based encoder was applied to to learn the correlation between multiple clauses in a document. They also encoded relative position and global predication information that they claim helped capture the causality between clauses. Recently, \citet{poria2020recognizing} extended the task of emotion-cause extraction for conversation and released a dataset called RECCON containing $1,000+$ dialogs accompanied by $~10,000$ emotion-cause pairs.

\begin{figure}[t]
    \centering
    \subfloat[Emotion flip reasoning]{
    \includegraphics[width=\columnwidth]{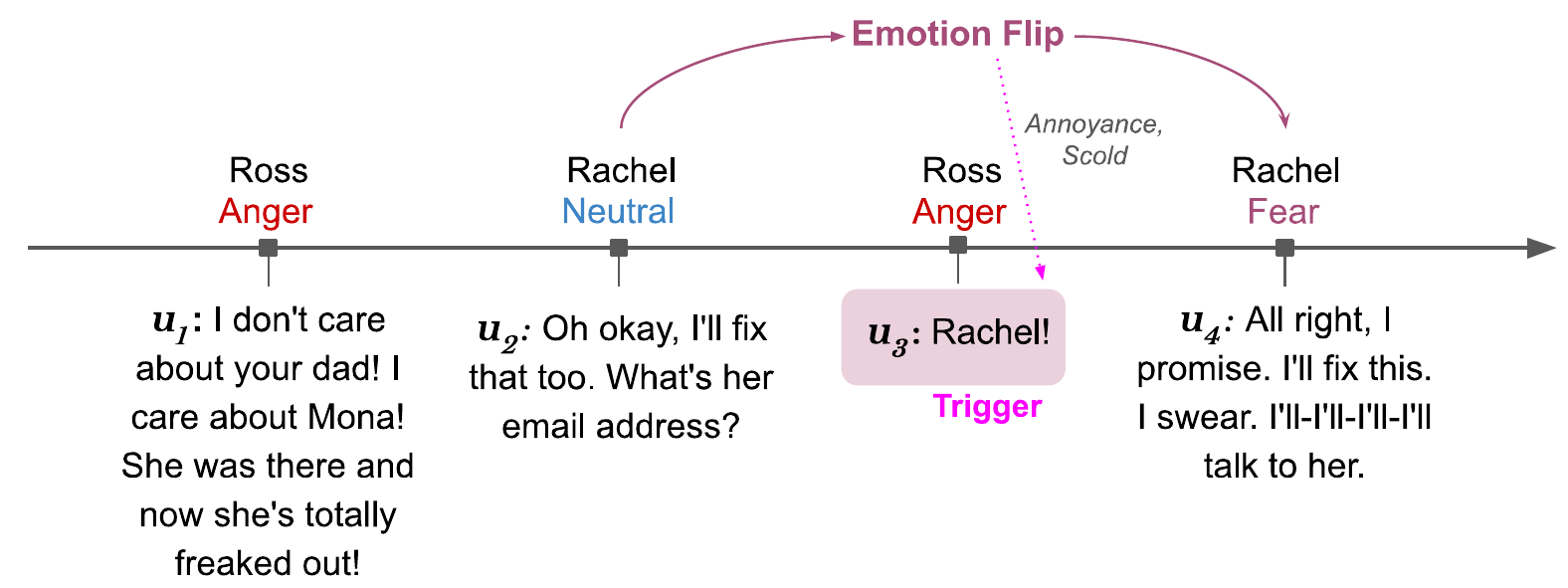}}\\
    \subfloat[Emotion-cause extraction]{
    \includegraphics[width=\columnwidth]{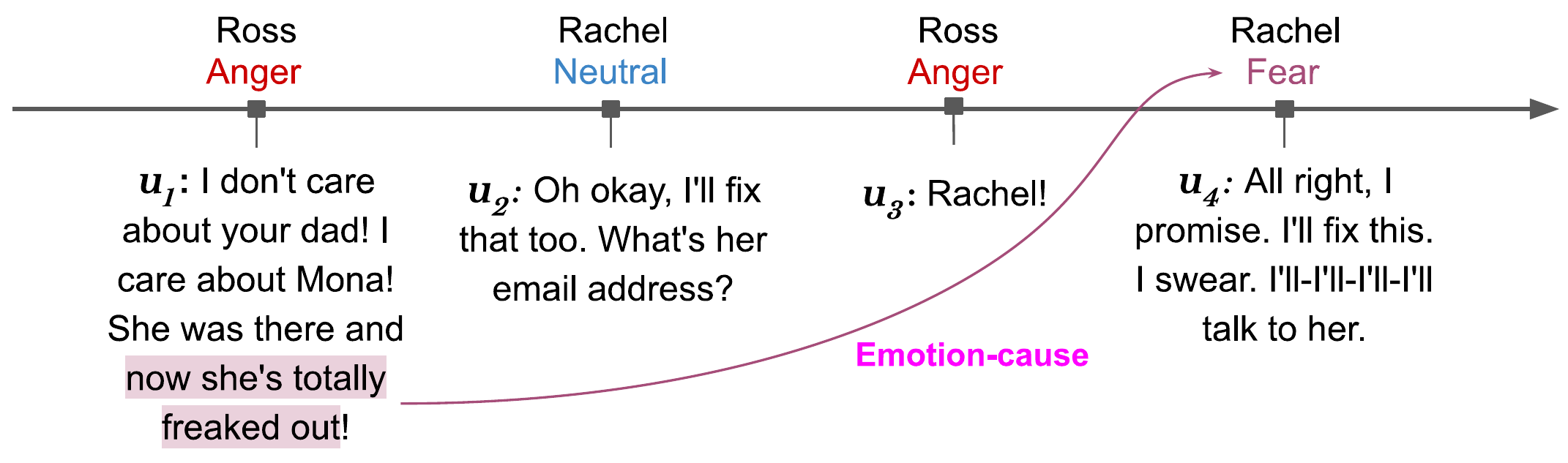}}
    \caption{A sample dialogue to illustrate the difference between emotion-cause extraction in conversation and emotion-flip reasoning.}
    \label{fig:eg_ece_efr}
    \vspace{-6mm}
\end{figure}

{\textbf{How is our task different?}} EFR represents a novel paradigm in NLP as it deals explicitly and quantitatively with identifying emotion instigators. While word embeddings may contain some implicit information about utterance meaning and emotion dynamics, they provide no explainability for an emotion-flip, and hence, cannot be used as a potential feedback mechanism to a response generator. Additionally, the two tasks, namely emotion-cause extraction in conversation and emotion-flip reasoning (EFR), may seem similar at an abstract level; nonetheless, they differ considerably at the surface level. While emotion-cause extraction in conversation aims to extract a text span that acts as grounds for the elicited emotion, EFR is a more speaker-specific task that highlights the instigators responsible for a ``flip" in the speakers' emotion. In our case, the instigators (or causes) for an emotion flip come from a finite set of predefined labels, in contrast with the infinite possibilities of a span that the emotion-cause extraction task can extract. In order to reinforce the difference between the two tasks, we show a sample dialogue in Figure \ref{fig:eg_ece_efr} from MELD-I with annotated EFR and emotion-cause labels. It can be observed that the reason behind the emotion \textit{fear} in utterance $u_4$ comes from utterances $u_1$ and $u_3$. On the other hand, the emotion flip from \textit{neutral} to \textit{fear} (from utterance $u_2$ to $u_4$) was triggered by the utterance $u_3$ because of the feelings of \textit{annoyance} and \textit{scold} being instigated in the target speaker.

\section{Dataset}
    \label{sec:dataset}
    \label{sec:dataset}

In this work, we use the existing MELD dataset \cite{meld} and augment it with EFR labels to get the new \textbf{MELD-I} (abbreviation of MELD-Instigators) dataset. MELD is a chit-chat dialogue dataset compiled from a famous TV series, F.R.I.E.N.D.S.\footnote{\url{https://www.imdb.com/title/tt0108778/}} It contains $1,433$  multi-party dialogues with $13,708$ utterances. Each utterance  is associated with one of the six Ekman's \cite{ekman} emotion labels -- \textit{anger}, \textit{fear}, \textit{disgust}, \textit{sadness}, \textit{joy}, and \textit{surprise}. Additionally, the absence of emotion is marked with the \textit{neutral} label. We include new ground-truth labels for EFR to get MELD-I. 
{We call the proposed MELD-I as a ``new" dataset primarily due to the \textit{\textbf{manually annotated}} instigator labels. Specifically, we propose new labels and convert the original MELD dataset into instances based on emotion-flips as illustrated in Table III  resulting in $1161$ dialogue instances. We manually annotate all these instances with the proposed instigator labels. However, since we use the dialogues and emotion labels from the MELD dataset, we keep the name of the new dataset derived from it- MELD-I.}

\subsection{Instigator Labels}
\label{sec:inst_labs}
To understand the emotional dynamics of the speakers in a conversation, it is imperative to reason out any change/flip of the emotion of any speaker. Following the Cognitive Appraisal Theory by Lazarus et al. \cite{lazarus}, which states that emotions are a result of appraisals, we aim to identify these appraisals for each emotion flip in the dialogue. These instigators follow the following properties:
\begin{itemize}
    \item The instigators need not be unique to an emotion flip. For example, \textit{threat} can instigate the emotion flip \textit{joy} $\rightarrow$ \textit{fear} as well as the flip \textit{anger} $\rightarrow$ \textit{fear}.
    \item An emotion flip need not necessarily arise from the same set of instigators. For example, the emotion flip \textit{neutral} $\rightarrow$ \textit{fear} can be caused by \textit{threat} and \textit{challenge} in different situations.
    \item There can be more than one instigator corresponding to a single emotion flip. For example, for the emotion flip \textit{neutral} $\rightarrow$ \textit{fear}, the instigator can be both \textit{threat} and \textit{challenge}.
    \item The instigators cannot be emotions themselves. For example, for the emotion flip \textit{neutral} $\rightarrow$ \textit{surprise}, the instigator cannot be \textit{joy}.
\end{itemize}

We organize these instigators in a 2-level hierarchy. The first level presents a coarser representation of the instigators with $14$ labels, while the second level defines all $27$ instigators as fine-grained representation. We present the hierarchy of instigators and their definitions in the supplementary.
Further, these instigators can be divided into three sets, based on the target emotion they can instigate- positive, negative, and neutral. A detailed discussion can be found in the supplementary.

\subsection{Annotation Process}
\label{sec:annotaion}
The first step in our annotation process is the instance creation for each emotion flip, followed by the trigger identification and instigator labeling. Table \ref{tab:data_creation2} presents the outcome of the annotation process for the example shown in Figure \ref{fig:data_example2}. We explain these steps in detail below.

\begin{enumerate}[leftmargin=*]
    \item \textbf{Instance creation:} For each emotion flip of a speaker, we create a new instance. The instance contains the utterances from the beginning of the dialogue till the target utterance (emotion flipped utterance). Utterances $u_3$ and $u_5$ are the target utterances in Table \ref{tab:data_creation2}, and utterances $\langle u_1, u_2, u_3\rangle$  and  $\langle u_1, u_2, u_3, u_4, u_5\rangle$  are the respective candidate triggers for the target utterances.
    Among these candidates, $u_2,u_3$ are the triggers for the target $u_3$, while utterance $u_4$ instigates the emotion flip in the target $u_5$. Intuitively, the last utterance of each instance is the target utterance -- the location of emotion flip. Note that we remove all such dialogues from MELD that do not contain an emotion flip which removed $271$ dialogues from the set.
    \item \textbf{Trigger identification:} After creating an instance for each emotion flip, we identify a set of trigger utterances that cause the emotion to flip at the target. We mark each utterance that acts as a trigger as `Yes' and the ones not contributing as `No'. The two instances in Table \ref{tab:data_creation2} have utterances $\langle u_2,u_3\rangle$ and $\langle u_4\rangle$ as triggers for the target utterances $u_3$ and $u_5$, respectively.
    \item \textbf{Instigator labeling:} Finally, we assign one or more instigator labels to each trigger utterance corresponding to the target utterance. For example, as presented in Table \ref{tab:data_creation2}, we assign `\textit{nervousness}' and `\textit{adoration}' instigators to the trigger utterances $u_2$ and $u_3$, respectively, for the target $u_3$. Similarly, for the target utterance $u_5$ in Table \ref{tab:data_creation2}, we annotate the trigger $u_4$ with two instigator labels `\textit{annoyance}' and `\textit{challenge}'. It is evident that the instigator identification is a multi-label problem.
\end{enumerate}

\begin{table*}[h!]
\centering
\caption{Dataset development for an instance shown in Figure \ref{fig:data_example2}.  {(a)}  Original dialogue from MELD; {(b,d)}  Two instances corresponding to the two emotion flips in (a); (c,e)  Trigger and instigator annotations for both instances.
}
\subfloat[An example dialogue from MELD.]{
\resizebox{0.7\textwidth}{!}{%
\begin{tabular}{|l|l|p{25em}|l|l}
    \cline{1-4} 
     & \bf Speaker & \bf Utterance & \bf Emotion & \\ \cline{1-4}
     $u_1$ & Ross & No! No sir umm, she means a lot to me. I mean, I care I-I love Rachel. & Fear & \\ \cline{1-4}
     $u_2$ & Mona & What?! & Surprise \\ \cline{1-4}
     $u_3$ & Ross & Oh but not that way. I mean I mean I’m not in love with her. I love her like a, like a friend. & Joy & \cellcolor{LightCyan} Emotion flip - 1 \\ \cline{1-4}
     $u_4$ & Dr. Green & Oh really? That’s how you treat a friend? You get her in trouble and then refuse to marry her? & Anger & \\ \cline{1-4}
     $u_5$ & Ross & Hey! I offered to marry her! & Anger & \cellcolor{LightCyan} Emotion flip - 2 \\ \cline{1-4}
 \end{tabular}
}
}
\\
\subfloat[Instance - 1.]{
\resizebox{0.48\textwidth}{!}{%
\begin{tabular}{|l|l|p{20em}|l|}
    \hline 
     & \bf Speaker & \bf Utterance & \bf Emotion \\ \hline
     $u_1$ & Ross & No! No sir umm, she means a lot to me. I mean, I care I-I love Rachel. & Fear \\ \hline
     $u_2$ & Mona & What?! & Surprise \\ \hline
     \rowcolor{LightCyan} $u_3$ & Ross & Oh but not that way. I mean I mean I’m not in love with her. I love her like a, like a friend. & Joy \\ \hline
 \end{tabular}}
}
\subfloat[MELD-I Annotation: Trigger/Instigator.]{
\resizebox{0.48\textwidth}{!}{%
\begin{tabular}{|l|l|p{18em}|l|l|p{5em}|}
    \hline 
     & \bf Speaker & \bf Utterance & \bf Emotion & \bf Trigger & \bf Instigator \\ \hline
     $u_1$ & Ross & No! No sir umm, she means a lot to me. I mean, I care I-I love Rachel. & Fear & No & - \\ \hline
     $u_2$ & Mona & What?! & Surprise & Yes & Nervousness \\ \hline
     \rowcolor{LightCyan} $u_3$ & Ross & Oh but not that way. I mean I mean I’m not in love with her. I love her like a, like a friend. & Joy & Yes & Adoration \\ \hline
\end{tabular}}
}
\\
\subfloat[Instance - 2.]{
\resizebox{0.48\textwidth}{!}{%
\begin{tabular}{|l|l|p{20em}|l|}
    \hline 
     & \bf Speaker & \bf Utterance & \bf Emotion \\ \hline
     $u_1$ & Ross & No! No sir umm, she means a lot to me. I mean, I care I-I love Rachel. & Fear \\ \hline
     $u_2$ & Mona & What?! & Surprise \\ \hline
     $u_3$ & Ross & Oh but not that way. I mean I mean I’m not in love with her. I love her like a, like a friend. & Joy \\ \hline
     $u_4$ & Dr. Green & Oh really? That’s how you treat a friend? You get her in trouble and then refuse to marry her? & Anger \\ \hline
     \rowcolor{LightCyan} $u_5$ & Ross & Hey! I offered to marry her! & Anger \\ \hline
 \end{tabular}
}}
\subfloat[MELD-I Annotation: Trigger/Instigator.]{
\resizebox{0.48\textwidth}{!}{%
\begin{tabular}{|l|l|p{20em}|l|l|p{5em}|}
    \hline 
     & \bf Speaker & \bf Utterance & \bf Emotion & \bf Trigger & \bf Instigator \\ \hline
     $u_1$ & Ross & No! No sir umm, she means a lot to me. I mean, I care I-I love Rachel. & Fear & No & - \\ \hline
     $u_2$ & Mona & What?! & Surprise & No & - \\ \hline
     $u_3$ & Ross & Oh but not that way. I mean I mean I’m not in love with her. I love her like a, like a friend. & Joy & No & - \\ \hline
     $u_4$ & Dr. Green & Oh really? That’s how you treat a friend? You get her in trouble and then refuse to marry her? & Anger & Yes & Annoyance, Challenge \\ \hline
     \rowcolor{LightCyan} $u_5$ & Ross & Hey! I offered to marry her! & Anger & No & - \\ \hline
\end{tabular}}
}
\label{tab:data_creation2}
\vspace{-6mm}
\end{table*}

We employ the services of three annotators\footnote{They are NLP researchers or linguistics by profession; their age ranges between $30-45$ years.} to annotate MELD-I -- two of them in the first stage of annotation, while the service of the third expert was sought to resolve any disagreement. We compute Krippendorff's Alpha inter-annotator agreement  \cite{krippendorff2011computing} to measure the consistency in the annotation. For trigger identification, we obtain the inter-annotator agreement between annotators A and B as $\alpha_{AB} = 0.817$, between annotators B and C as $\alpha_{BC} = 0.820$, and between annotators A and C as $\alpha_{AC} = 0.811$. 
We take the average of these three to get the overall agreement rating, i.e., $\alpha = 0.816$.
For the instigator annotation, $\alpha_{AB} = 0.511$, $\alpha_{BC} = 0.545$, and $\alpha_{AC} = 0.540$. The average agreement comes out to be $\alpha = 0.532$. The low value for the latter is attributed to the multi-label characteristic of the task.
\subsection{Dataset Statistics}
\label{sec:dataset_stats}
\begin{figure}[h!]
    \centering
    \subfloat[Coarse-grained \label{fig:inst_dist14}]{
    \includegraphics[width=0.48\columnwidth]{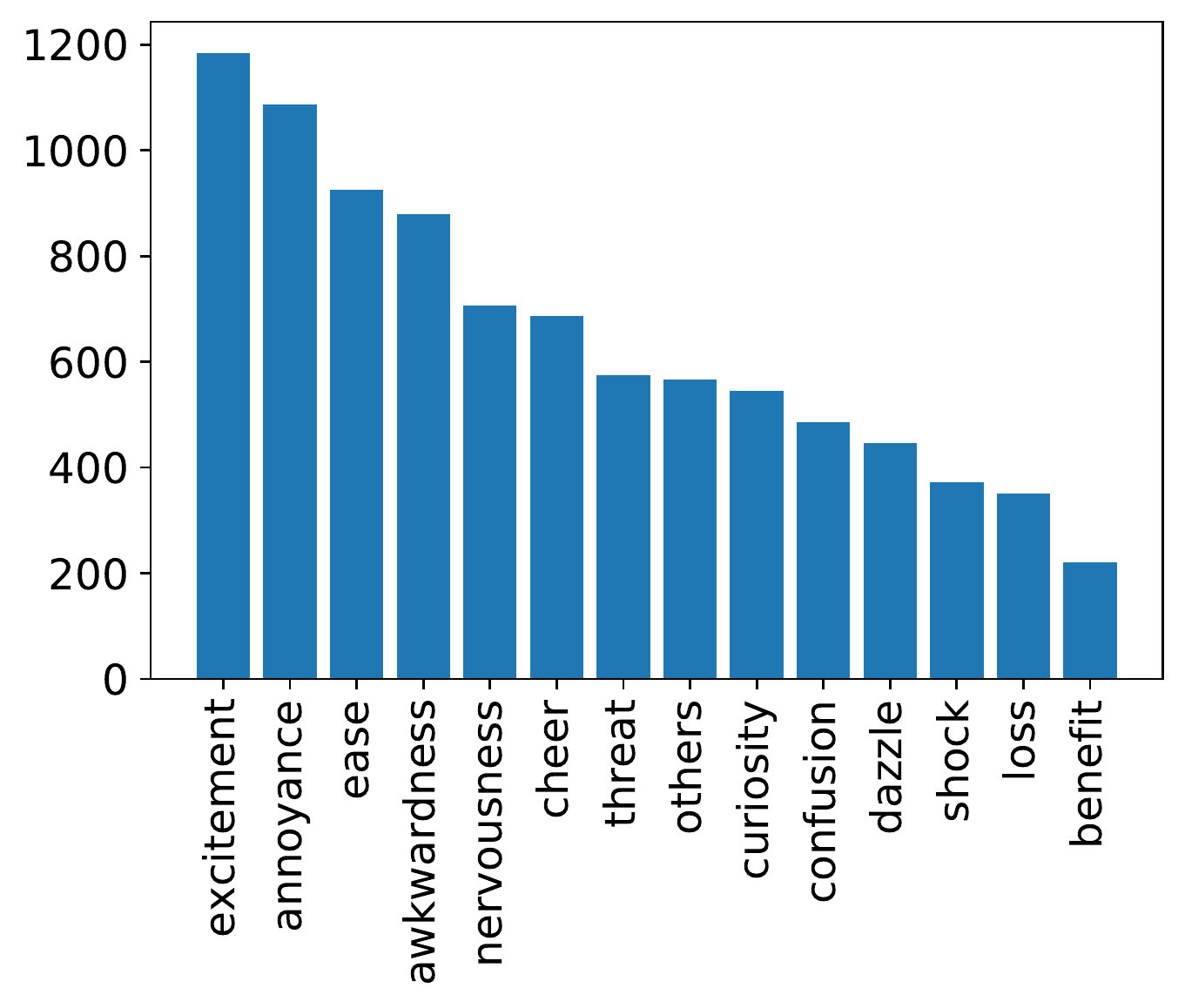}}
    \subfloat[Fine-grained \label{fig:inst_dist27}]{
    \includegraphics[width=0.48\columnwidth]{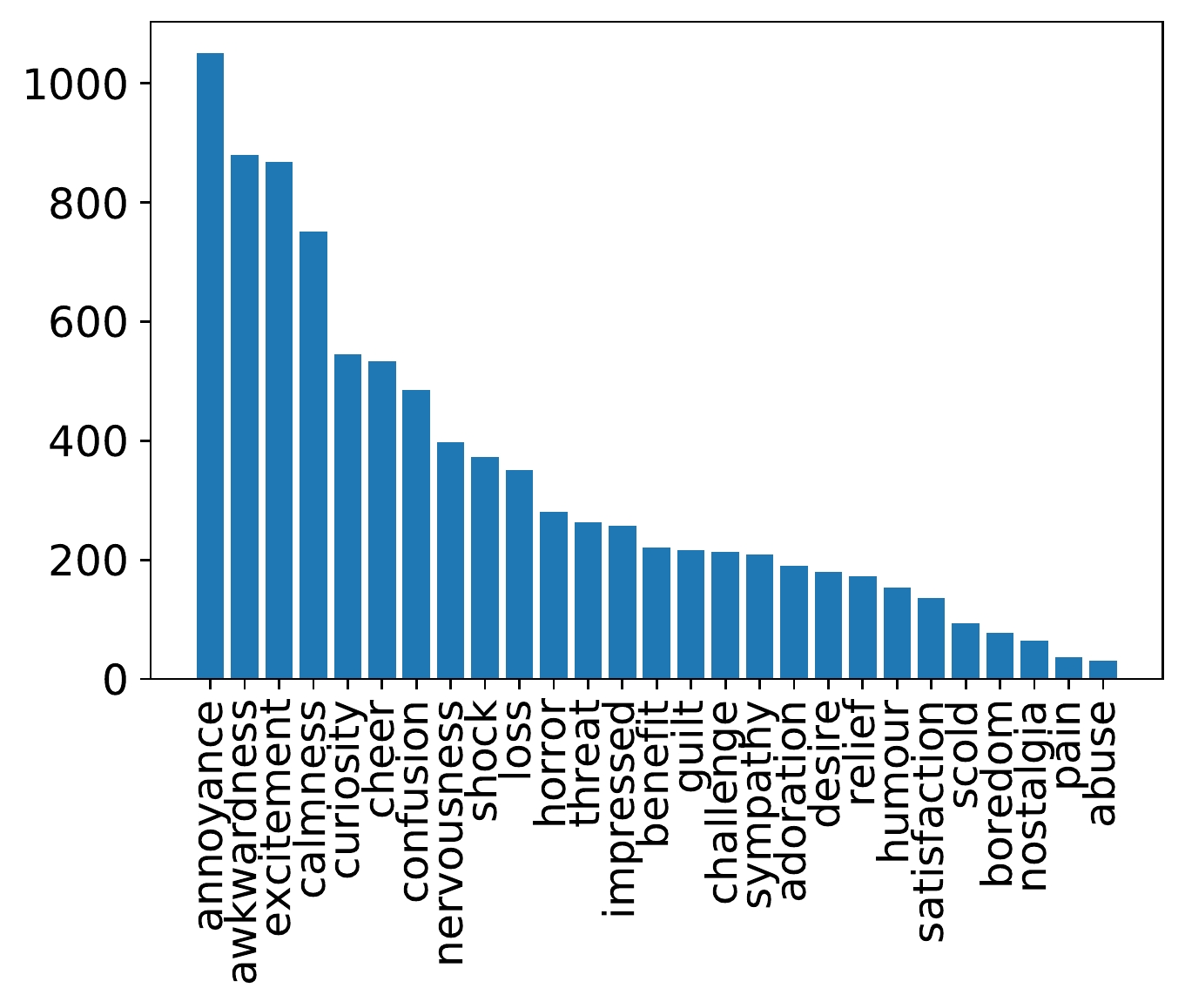}}
      \caption{Distribution of EFR instigators in MELD-I.}\label{fig:inst_dist}
    \vspace{-5mm}
\end{figure}

We show a brief statistic of  MELD-I  in Table \ref{tab:dataset} along with the distribution of emotion flips. We also show the distribution of instigators in Figure \ref{fig:inst_dist}. We can observe from Figure \ref{fig:inst_dist27} that the distribution of fine-grained instigator labels is skewed towards a few of the instigators. As an attempt to reduce the skewness, we group similar instigator labels and obtain a reduced set of $14$ instigators in the coarse-grained setup (c.f. Figure \ref{fig:inst_dist14}).

It is interesting to note that emotion flips can be divided into two categories -- {\bf positive} emotion flips (\{\textit{anger}, \textit{fear}, \textit{disgust}, \textit{sadness}\} $\rightarrow$ \{\textit{joy}, \textit{surprise}, \textit{neutral}\}) and {\bf negative} emotion flips (\{\textit{joy}, \textit{surprise}\} $\rightarrow$ \{\textit{anger}, \textit{fear}, \textit{disgust}, \textit{sadness}, \textit{neutral}\}).
The flips \{\textit{neutral}\} $\rightarrow$ \{\textit{joy}, \textit{surprise}\} are also considered as positive emotion flips whereas \{\textit{neutral}\} $\rightarrow$ \{\textit{anger}, \textit{fear}, \textit{disgust}, \textit{sadness}\} are considered as negative emotion flips. 
Considering the above categorization of emotion flips, we observe that not all instigators can result in all emotions flips. For example, it is improbable for a person to feel \textit{joy} because of \textit{guilt} -- for an emotion flip with the target emotion \textit{joy}, the instigator can almost never be \textit{guilt}. Our observation of the annotated dataset is in line with this phenomenon.

Consequently, we divide our instigator labels into three sets -- positive, negative, and ambiguous (c.f. Section \ref{sec:app_direction}). 
We observe that for a positive emotion flip, only the instigators belonging to the positive and ambiguous set of instigators are responsible. Similarly, for a negative emotion flip, the negative and ambiguous sets are applicable.
    
\section{Methodology}
    \label{sec:methodology}
    \begin{figure*}[h!]
    \centering
    \includegraphics[width=\textwidth]{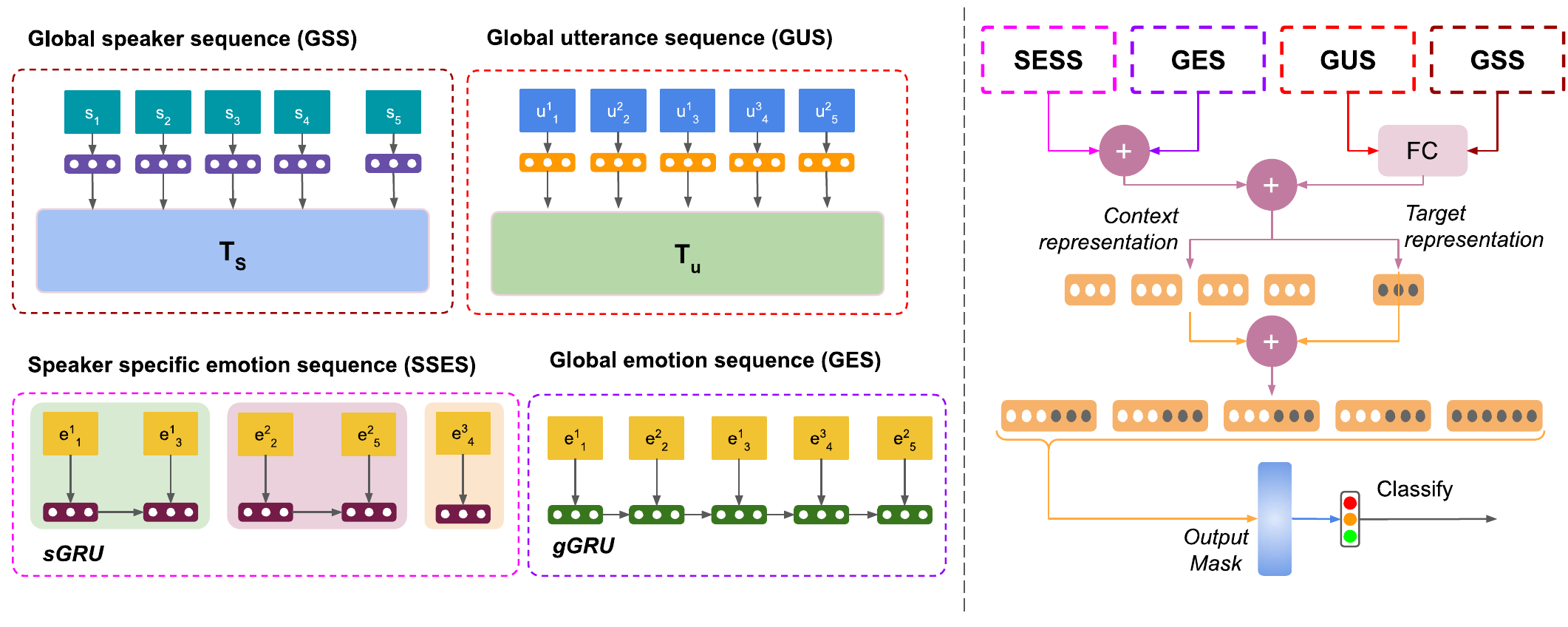}
    \caption{The proposed \name\ architecture. Input: {$\{\langle u_1,s_1,e_1 \rangle$, $\langle u_2,s_2,e_2 \rangle$, $\langle u_3,s_1,e_3 \rangle$, $\langle u_4,s_3,e_4 \rangle$, $\langle u_5,s_2,e_5 \rangle\}$}, where $\langle u_i,s_j,e_i \rangle$ represents the utterance $u_i$ by a speaker $s_j$ and its associated emotion $e_i$. Target (emotion flipped) utterance: $\langle u_5,s_2,e_5 \rangle$ as $e_2 \neq e_5$.
    }
    \label{fig:model}
    \vspace{-7mm}
\end{figure*}

This section dwells on our proposed model, \name, to identify the EFR instigator labels for each emotion flip. The instigator identification task is a multi-label instance classification problem, as more than one instigator is possible for each trigger. \name\ models the global utterance sequence (\textit{aka.} dialogue context) and speaker dynamics to capture the underlying semantics in the dialogue. Moreover, considering the strong relationship of emotion with our task, we also encode the emotion sequence of the utterances in \name. In total, \name\ has four submodules that exploit the global and speaker-specific dialogue and emotion dynamics -- Global Utterance Sequence (GUS), Global Speaker Sequence (GSS), Global Emotion Sequence (GES), and Speaker-Specific Emotion Sequence (SSES). Finally, we combine the outputs of these four modules through a series of fully-connected layers followed by a 14/27 neurons sigmoid layer for both coarse-grained and fine-grained instigator identifications. Furthermore, at the penultimate layer, we apply an output mask to filter out the improbable instigator labels for the underlying emotion flip. The output mask assists the model in focusing on the probable labels and blocks the gradients for the unlikely labels to propagate back to the network. Below, we describe each module of \name\ in detail.  Figure \ref{fig:model} presents the architecture of \name.
\begin{table}[t]
\centering
\centering
\caption{Statistics of the dataset, MELD-I.}
    \subfloat[MELD-I dataset for Emotion Flip Reasoning (EFR)]{
\resizebox{\columnwidth}{!}
{
\begin{tabular}{|c|c|c|c|}
    \hline
         \bf Split & \bf \#Dialogue with Flip & \bf \#Utterance with Flip & \bf \#Triggers  \\ \hline 
         
         \hline 
         \bf Train & 834 & 4001 & 5262 \\
         \bf Dev & 95 & 427 & 495 \\
         \bf Test & 232 & 1002 &  1152  \\ \hline
    \end{tabular}}}
    
\subfloat[Frequency of emotion flips with respect to emotions]{
\resizebox{\columnwidth}{!}{%
\begin{tabular}{c|l|c:c:c:c:c:c:c|}
\multicolumn{2}{c}{} & \multicolumn{7}{c}{\textbf{Target}} \\
\cline{3-9}

\multicolumn{1}{c}{} & \multicolumn{1}{c}{} & \multicolumn{1}{|c|}{\textbf{Disgust}} & \multicolumn{1}{c|}{\textbf{Joy}} & \multicolumn{1}{c|}{\textbf{Surprise}} & \multicolumn{1}{c|}{\textbf{Anger}} & \multicolumn{1}{c|}{\textbf{Fear}} & \multicolumn{1}{c|}{\textbf{Neutral}} & \multicolumn{1}{c|}{\textbf{Sadness}} \\ \cline{2-9}

\multirow{8}{*}{\rotatebox{90}{\bf Source}} & \textbf{Disgust} & 0 & 24 & 30 & 47 & 6 & 76 & 13 \\ \cline{2-2} \cdashline{3-9}
\multicolumn{1}{c|}{} & \textbf{Joy} & 34 & 0 & 169 & 86 & 42 & 665 & 81 \\ \cline{2-2} \cdashline{3-9}
\multicolumn{1}{c|}{} & \textbf{Surprise} & 39 & 186 & 0 & 137 & 32 & 400 & 70 \\ \cline{2-2} \cdashline{3-9}
\multicolumn{1}{c|}{} & \textbf{Anger} & 37 & 96 & 104 & 0 & 20 & 318 & 99 \\ \cline{2-2} \cdashline{3-9}
\multicolumn{1}{c|}{} & \textbf{Fear} & 7 & 20 & 23 & 45 & 0 & 87 & 27 \\ \cline{2-2} \cdashline{3-9}
\multicolumn{1}{c|}{} & \textbf{Neutral} & 84 & 616 & 487 & 370 & 103 & 0 & 257 \\ \cline{2-2} \cdashline{3-9}
\multicolumn{1}{c|}{} & \textbf{Sadness} & 17 & 78 & 60 & 72 & 28 & 238 & 0 \\ \cline{2-9}
\end{tabular}%
}}
\label{tab:dataset}
\vspace{-5mm}
\end{table}

\paragraph{\bf Global Utterance Sequence (GUS)}
The principle information about a dialogue lies in the utterances spoken in it. Thus, we employ GUS to encode the utterance sequence. We use a Transformer \cite{transformer} encoder to extract a hidden representation $h_i^u$ for each utterance $u_i$. For each, utterance, $u_i^{s_i}$, in the dialogue, we get an encoded vector, $h_i^u$, after this state, i.e. $\forall u_i^{s_i}, h_i^u = T_u(u_i^{s_i})$. Thus, $h_i^u$ represents the context aware representation of the $i^{th}$ utterance of the dialogue $D$.

\paragraph{\bf Global Emotion Sequence (GES)} In this module, we employ a single-layer GRU \cite{chung2014empirical}, $gGRU$, to capture the global emotion sequence of the dialogue. We hypothesize that the knowledge of emotion sequence would assist the model in capturing a high-level snapshot of the emotion flow among speakers. We feed the emotion sequence of the dialogue, $\{e_1, e_2, ..., e_t\}$, as input to the GRU where each emotion $e_i$ is represented by a one-hot vector of dimension $7$. As a result, we obtain the hidden representation 
as follows: $[h_1^e,.., h_2^e,.., h_t^e] = gGRU(e_1, e_2, .., e_t)$.

\paragraph{\bf Speaker-Specific Emotion Sequence (SSES)} Each emotion flip is associated with a speaker. Thus, we hypothesize that the sequence of emotions at the speaker level is crucial and would exploit the emotion dynamics of each speaker considering the target speaker. Moreover, it would distinguish between the emotional states of the target speaker and other speakers. To achieve this, we employ separate GRUs for each speaker in an instance.

For example, if there are three distinct speakers in an instance (c.f. instance 2 in Table \ref{tab:data_creation2}), we learn three GRUs. 
For each speaker, we extract its emotion from the dialogue and create the input for GRUs as follows. Let an instance with five utterances of three distinct speakers be given as $\{\langle u_1,s_1,e_1 \rangle$, $\langle u_2,s_2,e_2 \rangle$, $\langle u_3,s_1,e_3 \rangle$, $\langle u_4,s_3,e_4 \rangle$, $\langle u_5,s_2,e_5 \rangle\}$, where $u_i$ and $e_i$ denote the utterance and associated emotion at turn $i$ by speaker $s_j$. We compile three inputs for each speaker as $\{e_1, e_3\}$, $\{e_2, e_5\}$, and $\{e_4\}$, and feed them to three speaker-specific GRUs (sGRU).
\begin{align}
    [\hat{h}_1^e, \hat{h}_3^e] & = sGRU_1(e_1,e_3) \nonumber \\
    [\hat{h}_2^e, \hat{h}_5^e] & = sGRU_2(e_2,e_5) \nonumber \\
    [\hat{h}_4^e] & = sGRU_3(e_4) \nonumber
\end{align}

Finally, we combine the hidden representations of GRUs by arranging them in the dialogue order for further processing, i.e.,
$\hat{H} = [\hat{h}_1^e, \hat{h}_2^e, \hat{h}_3^e, \hat{h}_4^e, \hat{h}_5^e]$.

\paragraph{\bf Global Speaker Sequence (GSS)} To explicitly capture the speaker information, their reactions with respect to other speakers, and their relationships, we also propose to encode the speaker sequence in \name. To capture the different reactions of a speaker with respect to the utterance of other speakers, we capture the speaker sequence by employing another Transformer encoder, $T_s$, which takes as input the sequence of speakers where each speaker is represented by a one-hot encoded vector. Each speaker goes through the Transformer encoder to give a speaker sequence aware representation, $h_i^s$, i.e. $\forall s_i, h_i^s = T_s(s_i)$. After this, we have a speaker sequence aware representation for each speaker, $h_i^s$, in the dialogue.

\paragraph{\bf Fusion} We fuse the outputs of the above four submodules in two steps. In the first step, we combine the dialogue-level utterance and speaker sequence to obtain a global view of the conversation through a fully-connected layer. In parallel, we combine the dialogue and speaker-level emotion dynamics to get the essence of the flow of emotions in the conversation. Subsequently, in the second step, we concatenate the two representations. As the effect of an utterance on the final emotion changes with the change of the target utterance, we append the target representation to each utterance before feeding it to the output layer for prediction. We can justify the appending operation through the example shown in Table \ref{tab:data_creation2}. It can be observed that utterance $u_2$ is present in both instances; however, it is the trigger only in the first instance. Moreover, in Figure \ref{fig:data_example1}, the same trigger utterance $u_3$ resulted in an emotion flip from \textit{neutral} to \textit{sadness} in the first instance, while it causes the speaker's emotion change to \textit{surprise} from \textit{joy} for the second instance. Finally, we apply gradient masking for the improbable instigators. 

\section{Experiments and Results}
    \label{sec:results}
    We perform experiments for both the granular levels -- coarse-grained and fine-grained. In the fine-grained setup, we observe a few instigator labels with a very low count. Since these labels are few in number, the model does not have sufficient evidence to learn a mapping from the input to such labels. Consequently, we compile another coarse-grained setup where we merge all instigator labels with count $<250$ into a set, called `\textit{other}'.  As a result, in total, we have three setups -- one fine-grained with $27$ instigator labels) and two coarse-grained (definition-based (c.f. Section \ref{sec:dataset}) and count-based) with $14$ instigator labels each. In all three setups, we employ sigmoid neurons with focal loss \cite{focal:Lin:2017:ICCV} for multi-label classification. We select the traditional precision, recall, and F1-score as our metrics of choice thus ensuring that our evaluation is consistent with existing practices and establishes a universal benchmark.

\subsection{Development Phase}
To find the best configuration for \name, we investigate the effect of each module in the development phase. We start with the GUS module as the backbone network and subsequently introduce other modules (GES, GSS, and SESS) in an incremental fashion. Table \ref{tab:ablation_modules} illustrates the results we obtain. Looking at the fine-grained setup, we notice a performance increase of $1.6\%$ in weighted F1 when we add the GES module to the backbone model. This performance increase is coherent with our argument that the inclusion of emotional information will help the model in learning a better mapping function. Additionally, the incorporation of speaker-specific modules (GSS and SESS) gives a performance boost of $0.6\%$ and $1.6\%$, respectively. We use the model consisting of all four submodules as our final architecture since it yields the best performance on the development set. After fine-tuning the hyperparameters during the development phase, we fix the configuration and evaluate \name\ on the test set.
\begin{table}[h!]
\centering
\caption{Results (W-F1) of fine-tuning on the development set. It shows the effect of each module when incorporated in \name. We obtain the best results when all four submodules are employed (last row).}
\resizebox{\columnwidth}{!}{%
\begin{tabular}{|l|c|c|c|}
\hline
\multirow{2}{*}{\textbf{Model}} & \multicolumn{2}{c|}{\textbf{Coarse-grained}} & \multirow{2}{*}{\textbf{Fine-grained}} \\ \cline{2-3} 

& \textbf{Defn-based} & \textbf{Count-based} & \\ \hline

\hline
\textbf{GUS} & 41.4 & 37.0 & 29.9 \\ \hdashline
\textbf{+ GES} & 41.9 & 37.1 & 31.5 \\ \hdashline
$\quad$\textbf{+ GSS} & 42.1 & 37.8 & 32.1 \\ \hdashline
$\quad \quad$\textbf{+ SESS (\name)} & \textbf{42.7} & \textbf{38.5} & \textbf{33.1} \\ \hline
\end{tabular}%
}
\label{tab:ablation_modules}
\vspace{-3mm}
\end{table}

\subsection{Baselines and Comparative Study}
Since the problem of instigator classification for EFR is novel, we adapt various related existing systems for comparison. Note that all these systems are recent state-of-the-arts and designed especially for the task of emotion recognition in conversation (ERC).

\begin{enumerate}[leftmargin=*]
    \item \textbf{DialogueGCN} \cite{dialoguegcn}: 
    It exploits  self and inter-speaker dependency to recognize emotion in conversations. It captures utterance embeddings with GRUs, followed by a graph convolution network (GCN) to leverage the speaker-level context, which is finally used for the task of emotion classification. 
    
    \item \textbf{AGHMN} \cite{aghmn}: It uses an attention-based GRU  to monitor the flow of information through a hierarchical memory network. The attention weights are calculated over the contextual utterances in the conversation and combined for the final classification.
    
    \item \textbf{TL-ERC} \cite{tlerc}: It is a transfer learning-based framework for emotion recognition in conversation. The weights of a hierarchical encoder-decoder model, trained for the dialogue generation task, are used for emotion classification.
    
    \item \textbf{DialogXL} \cite{dialogxl}: It
    modifies XLNet and evaluates it for emotion recognition in conversation (ERC). The authors changed the segment-level recurrence mechanism to an utterance-level recurrence mechanism so that XLNet could be mapped to a dialogue setting and incorporated dialogue-aware self-attention to capture the intra- and inter-speaker dependencies in a conversation.
    
    \item \textbf{BERT} \cite{devlin2018bert}: 
    It is basically an encoder stack of transformer architecture \cite{vaswani2017attention}.
\end{enumerate}

\begin{table}[t]
\centering
\caption{Comparative results on coarse-grained and fine-grained instigators. All the metrics are  weighted average over all instigator classes.}
\resizebox{\columnwidth}{!}{%
\begin{tabular}{|l|c|c|c|c|c|c|c|c|c|}
\hline
\multirow{3}{*}{\textbf{Model}} & \multicolumn{6}{c|}{\textbf{Coarse-grained}} & \multicolumn{3}{c|}{\multirow{2}{*}{\textbf{Fine-grained}}} \\ \cline{2-7} 

& \multicolumn{3}{c|}{\textbf{Defn-based}} & \multicolumn{3}{c|}{\textbf{Count-based}} & \multicolumn{3}{c|}{}\\ \cline{2-10} 
 & \textbf{Pre} & \textbf{Rec} & \textbf{F1} & \textbf{Pre} & \textbf{Rec} & \textbf{F1} & \textbf{Pre} & \textbf{Rec} & \textbf{F1} \\ \hline
 
 \hline
\textbf{AGHMN} & 7.6 & 20.4 & 11.07 & 8.5 & 25.4 & 12.73 & 15.1 & 17.6 & 16.3 \\ \hdashline
\textbf{TL-ERC} & 9.6 & 49.0 & 16.6 & 14.4 & 54.8 & 21.7 & 7.1 & 33.0 & 12.8 \\ \hdashline
\textbf{DGCN} & 12.5 & 67.0 & 19.8 & 18.5 & 70.2 & 27.5 & 10.5 & 67.2 & 17.5 \\ \hdashline
\textbf{DialogXL} & 7.3 & 37.5 & 12.22 & 8.8 & 43.7 & 14.64 & 9.8 & 34.2 & 15.3 \\ \hdashline
\textbf{BERT} & 18.3 & 62.9 & 27.2 & 17.5 & 59.2 & 26.3 & 14.8 & 55.1 & 21.7 \\ \hline

\hline
\textbf{\name} & \textbf{24.3} & 58.6 & \textbf{31.6} & \textbf{28.3} & 63.4 & \textbf{37.5} & \textbf{26.5} & 55.6 & \textbf{33.3} \\ \hline
\end{tabular}%
}
\label{tab:results}
\end{table}
\begin{table*}[ht]
\centering
\caption{Fine-grained analysis: Actual and predicted instigator labels for an EFR instance. \name\ predicts one and two correct instigator labels for the two trigger utterances, $u_4$ and $u_5$, respectively. In each case, it wrongly predicts one instigator. In contrast, BERT reports a high percentage of \textit{false positives}.
}
\resizebox{\textwidth}{!}{%
\begin{tabular}{|l|l|p{15em}|c|c|c|c|l|}
\hline
&\multirow{3}{*}{\textbf{Speaker}} & \multirow{3}{*}{\textbf{Utterance}} & \multirow{3}{*}{\textbf{Emotion}} & \multirow{3}{*}{\textbf{Trigger}} & \multicolumn{3}{c|}{\textbf{Instigator}} \\ \cline{6-8}
 & & & & & \multirow{2}{*}{\textbf{Gold}} & \multicolumn{2}{c|}{\textbf{Prediction}} \\ \cline{7-8}
 & & & & & & \multicolumn{1}{|c|}{\textbf{\name}} & \multicolumn{1}{|c|}{\textbf{BERT}}  \\ \hline

\hline
$u_1$ & Monica & Yeah, but without the costumes. & neutral & No & - & - & - \\ \hline
$u_2$ & Phoebe & Oh. & neutral & No & - & - & - \\ \hline
$u_3$ & Joey & And it's not fake, it's totally brutal. & neutral & No & - & - & - \\ \hline
$u_4$ & Chandler & Yeah, it's two guys in a ring, and the rules are: They are no rules. & neutral & Yes & \multirow{2}{5em}{confusion, curiosity} & \multirow{2}{7em}{confusion, {\color[HTML]{FE0000} shock}} &  {\color[HTML]{FE0000} excitement}, {\color[HTML]{FE0000} nervousness},  {\color[HTML]{FE0000} shock}\\ \hline
$u_5$ & Monica & So you can like, bite, and pull people's hair and stuff? & surprise & Yes & \multirow{2}{5em}{confusion, shock} & \multirow{2}{7em}{confusion, shock, {\color[HTML]{FE0000} curiosity}} & {\color[HTML]{FE0000} curiosity}, shock \\ \hline
\end{tabular}%
}
\label{tab:error27}
\vspace{-3mm}
\end{table*}

Similar to \name, we perform instance-wise experiments with output masking for each baseline. {That is, all the improbable instigators
are masked. Moreover, since we provide emotion labels as input to our model, we do the same with the baselines.}

Table \ref{tab:results} shows that \name\ outperforms all baselines with reasonable margin across all setups. In the definition-based coarse-grained setup, we obtain  $11.07\%, 16.6\%, 19.8\%,$ $12.2\%$, and $27.2\%$ weighted-F1 for AGHMN, TL-ERC, DGCN, DialogXl, and BERT respectively. In comparison, \name\ yields $31.6\%$ W-F1 in the same setup -- an increment of $4.4$ points over the best performing baseline (BERT). We observe a similar trend for the count-based coarse-grained setup with \name\ and the best baseline (DGCN) reporting $37.5\%$ and $27.5\%$ W-F1, respectively -- a difference of $10$ points. In the fine-grained setup,  the performance of the baselines (ranging between $12.8\% - 21.7\%$) are significantly inferior to \name\ ($33.3\%$). It suggest that \name\ also accounts for the increase in instigator labels more efficiently than the existing baselines. \name\ beats all considered baselines in every setting but at the same time reports a weighted F1 score on the lower side indicating the difficulty of the problem statement.
\subsection{Result Analysis}
    As we can observe from the distribution of EFR instigators (c.f. Figure \ref{fig:inst_dist}), there is a significant label skewness. To inspect the learning of \name\ for individual labels, we analyze the results of top-3 (majority) and bottom-3 (minority) instigator labels w.r.t. the number of training instances in MELD-I. The top-3 labels are \textit{nervousness}, \textit{awkwardness}, and  \textit{excitement} in the coarse-grained setup, and \textit{annoyance}, \textit{awkwardness}, and  \textit{excitement} in the fine-grained setup. Similarly, The bottom-3 labels in the coarse-grained setup are \textit{shock}, \textit{dazzle}, and \textit{threat}, while instigators \textit{nostalgia}, \textit{pain}, and \textit{boredom} are the three least occurring labels in the fine-grained setup. 

Tables \ref{tab:classwise:results} report the results of \name\ and baselines for the majority and minority classes, respectively. As expected, the performance of each model for the majority classes is comparatively on the higher side of the spectrum than the performance on minority classes. Except for the instigators, \textit{nostalgia}, \textit{pain}, \textit{boredom} in the fine-grained minority cases and \textit{excitement} in the coarse-grained majority case, \name\ reports the best weighted-F1 for each case. The observed behaviour can be attributed to the fact that Transformer based architectures are data-hungry models, and thus they learn a better mapping for majority classes.

\begin{table}[t]
\centering
\caption{Class-wise comparative analysis (F1-score) for the top-3 (majority) and bottom-3 (minority) classes. $\langle$\textit{Ner: Nervousness}, \textit{Awk: Awkwardness}, \textit{Exc: Excitement}, \textit{Ann: Annoyance}, \textit{Shk: Shock}, \textit{Daz: Dazzle}, \textit{Tht: Threat}, \textit{Nos: Nostalgia}, \textit{Bor: Boredom}$\rangle$.}
\resizebox{\columnwidth}{!}{%
\begin{tabular}{|l|l|l|l|l|l|l|l|l|l|l|l|l|}
\hline
\multirow{3}{*}{\textbf{Model}} & \multicolumn{6}{c|}{\textbf{Top-3 Majority}} & \multicolumn{6}{c|}{\textbf{Bottom-3 Minority}} \\ \cline{2-13} 
 & \multicolumn{3}{c|}{\textbf{Coarse-grained}} & \multicolumn{3}{c|}{\textbf{Fine-grained}} & \multicolumn{3}{c|}{\textbf{Coarse-grained}} & \multicolumn{3}{c|}{\textbf{Fine-grained}} \\ \cline{2-13} 
 & \multicolumn{1}{c|}{\textbf{Ner}} & \multicolumn{1}{c|}{\textbf{Awk}} & \multicolumn{1}{c|}{\textbf{Exc}} & \multicolumn{1}{c|}{\textbf{Ann}} & \multicolumn{1}{c|}{\textbf{Awk}} & \multicolumn{1}{c|}{\textbf{Exc}} & \multicolumn{1}{c|}{\textbf{Shk}} & \multicolumn{1}{c|}{\textbf{Daz}} & \multicolumn{1}{c|}{\textbf{Tht}} & \multicolumn{1}{c|}{\textbf{Nos}} & \multicolumn{1}{c|}{\textbf{Pain}} & \multicolumn{1}{c|}{\textbf{Bor}} \\ \hline
\textbf{AGHMN} & 11.2 & 10.1 & 8.6 & 17.8 & 12.8 & 15.3 & 3.9 & 4.2 & 3.2 & \bf 13.4 & 7.2 & \bf 16.9 \\ \hdashline
\textbf{TL-ERC} & 23.0 & 23.3 & 18.8 & 14.9 & 23.5 & 14.6 & 2.2 & 10.6 & 5.4 & 6.6 & 0.6 & 1.8 \\ \hdashline
\textbf{DGCN} & 28.9 & 28.4 & 28.3 & 24.8 & 26.9 & 23.2 & 10.5 & 9.8 & 6.0 & 0.0 & \bf 7.6 & 2.1 \\ \hdashline
\textbf{DialogXL} & 12.2 & 14.7 & 12.6 & 13.9 & 11.9 & 11.1 & 2.1 & 3.6 & 4.2 & 6.4 & 5.2 & 2.9 \\ \hdashline
\textbf{BERT} & 36.0 & 26.8 & \bf 35.3 & 38.9 & 26.8 & 36.1 & 10.7 & 2.5 & 0.0 & 3.2 & 2.0 & 0.0 \\ \hline
\textbf{TGIF} & \bf 37.8 & \bf 35.7 & 28.4 & \bf 53.5 & \bf 35.8 & \bf 56.7 & \bf 35.1 & \bf 18.6 & \bf 9.8 & 12.5 & 5.5 & 0.0 \\ \hline
\end{tabular}%
}
\label{tab:classwise:results}
\vspace{-5mm}
\end{table}

\subsection{Qualitative Error Analysis}
In order to perform qualitative error analysis, we take a sample dialogue from our test set and show the gold and predicted labels in Table \ref{tab:error27} for the fine-grained setup. For the target utterance $u_5$, \name\ predicts \textit{confusion} and \textit{shock} instigators against the gold labels \textit{confusion} and \textit{curiosity} instigated by the trigger utterance $u_4$. Similarly, for the the trigger $u_5$, \name\ identifies two correct (\textit{confusion} and \textit{shock}) and one incorrect label (\textit{curiosity}). An abstract view of the prediction suffices that the set of instigator labels for the emotion flip target $u_5$ (without regarding the triggers separately) is same as the set of gold labels. On the other hand,  BERT (best baseline) commits many mistakes in both cases. It predicts one correct label for the trigger $u_5$ but no correct instigator for the trigger $u_4$. It can be observed that BERT gives precision scores of $0\%$ for the first trigger while a precision of $50\%$ is observed for the second trigger. Recall value also comes out to be $0\%$ and $50\%$ for the two triggers, respectively. In comparison, \name\ obtains moderate scores in both cases, i.e., recall $= 50.0\%$; precision $= 50.0\%$ in the first case and recall $= 100.0\%$; precision$= 66.7\%$ in the second case. A similar trend is observed for coarse-grained instigator labels. We note that, for a given target emotion, the BERT method consistently identifies the same instigators, giving little heed to the conversation context. In contrast, our approach takes both the target emotion and conversation context into account when identifying instigators, resulting in more accurate and nuanced predictions. Further, we show the zero-shot results of the proposed method in the supplementary.

\subsection{Directionality of Triggers}
\label{sec:app_direction}
In this work, we consider Ekman's emotion labels along with a label for no emotion (neutral). That is, we have six emotion labels, namely \textit{disgust}, \textit{joy}, \textit{surprise}, \textit{anger}, \textit{fear}, and \textit{sadness}. An emotion flip for a target speaker can occur between any two pair of emotions. In other words, we can have $42$ possible emotion flips in a dialogue. Based on the source-target emotion pairs, we analyse the effect of directionality of emotion flips in MELD-I. We show the frequency of emotion flips with respect to the source-target emotion pairs in Table \ref{tab:dataset}. Cell $(i,j)$ in the table represents the number of flips in MELD-I where the source emotion is $e_i$, and the resultant or target emotion is $e_j$. As discussed in Section \ref{sec:dataset_stats}, there can be two types of emotion flips -- \textit{positive} and \textit{negative}. Here, we see how the emotion flips frequency and instigators are dependant on the type of flips. Based on our ground-truth EFR labels, there are a total of $2612$ positive emotion flips and $2818$ negative emotion flips. Out of these flips, the flip \textit{neutral} $\rightarrow$ \textit{joy} is the most prominent positive emotion flip whereas the flip, \textit{joy} $\rightarrow$ \textit{neutral} is the most prominent negative emotion flip.

\begin{table}[t]
\centering
\caption{Result analysis on directionality of flips for coarse-grained and fine-grained instigators. All the metrics are  weighted average over all instigator classes.}
\resizebox{\columnwidth}{!}{%
\begin{tabular}{|l|c|c|c|c|c|c|c|c|c|}
\hline
\multirow{3}{*}{\textbf{Type of Flip}} & \multicolumn{6}{c|}{\textbf{Coarse-grained}} & \multicolumn{3}{c|}{\multirow{2}{*}{\textbf{Fine-grained}}} \\ \cline{2-7} 

& \multicolumn{3}{c|}{\textbf{Defn-based}} & \multicolumn{3}{c|}{\textbf{Count-based}} & \multicolumn{3}{c|}{}\\ \cline{2-10} 
 & \textbf{Pre} & \textbf{Rec} & \textbf{F1} & \textbf{Pre} & \textbf{Rec} & \textbf{F1} & \textbf{Pre} & \textbf{Rec} & \textbf{F1} \\ \hline
 
 \hline
\textbf{Negative to Positive} & \textbf{27.3} & 54.6 & 33.9 & 26.1 & 52.5 & 32.7 & 19.3 & 52.2 & 26.2 \\ \hdashline
\textbf{Positive to Negative} & 26.4 & \textbf{61.2} & \textbf{35.1} & \textbf{28.0} & \textbf{59.1} & \textbf{35.4} & \textbf{32.2} & \textbf{58.1} & \textbf{38.8} \\ \hline
\end{tabular}%
}
\label{tab:pos_neg_flips}
\vspace{-4mm}
\end{table}

Apart from the emotion flips which have opposite polarities at both ends, we can also have intra-polarity flips. For instance, flips like \textit{anger} $\rightarrow$ \textit{fear} is a negative to negative emotion flip, while \textit{joy} $\rightarrow$ \textit{surprise} is a positive to positive emotion flip. We see, for the intra-polarity cases, that the flip \textit{surprise} $\rightarrow$ \textit{joy} and the flip \textit{anger} $\rightarrow$ \textit{sadness} are the most prominent intra-positive and intra-negative flips, respectively. We also observe that most of the flips that result in a negative emotion (\textit{anger}, \textit{disgust}, \textit{fear}, \textit{sadness}) originate from \textit{joy}. On the other hand, the flips that result in a positive emotion (\textit{joy}, \textit{surprise}) originate from \textit{neutral}. We also observe that, for positive emotion flips, the top-3 frequent instigators are \textit{excitement}, \textit{cheer}, or \textit{impressed}. For negative emotion flip, \textit{awkwardness}, \textit{loss}, or \textit{annoyance} are the more frequent instigators.

In addition, we check the performance of our models on the two most prominent types of flip directions -- positive to negative and negative to positive. We show these results in Table \ref{tab:pos_neg_flips} and observe that positive to negative flips are better predicted by our model for all the classes of instigators. This can be attributed to the fact that our data contains more negative emotions, thus containing more negative instigators. Consequently, our model is able to learn those instigators in a better fashion. This result is encouraging as it is an indication that with more data, our model will be able to learn the instigators in a better way.

\subsection{Generalizability of \name}
In order to emphasise the relevance of EFR and evaluate the generalizability of the proposed methodology, \name, we perform a zero-shot experiment. We consider IEMOCAP \cite{iemocap} which consists of emotion annotated conversations on $16$ topics. We randomly sample $15$ conversations to construct emotion flip instances, with triggers identified as shown in Table III of the main text. We then task \name\ and BERT, the best baseline, with predicting the emotion flip instigators. After collecting the predictions, we asked $20$ human annotators to rate them based on correctness, completeness, and preference (\name\ vs. BERT). The cumulative results in Table \ref{tab:human_eval} indicate that while \name\ outperforms BERT, the latter is comparable in terms of completeness.

\begin{table}[h!]\centering
\caption{Human Evaluation Results on IEMOCAP \cite{iemocap} in a zero-shot setting. Scores are average across all evaluators.}
\begin{tabular}{lrrrr}\toprule
&Correctness &Completeness &Prefered Instigator set \\\midrule
BERT &2.67 &3.42 &25\% \\
\name &\textbf{3.21} &\textbf{3.44} &\textbf{75\%} \\
\bottomrule
\end{tabular}
\label{tab:human_eval}
\end{table}
\vspace{-6mm}
\section{Conclusion}
    \label{sec:conclusion}
    This paper focused on explaining the reasons for a flip of a speaker's emotion in a conversation. Our interest lies in revealing which cognitive appraisal instigated the flip. We introduced Emotion Flip Reasoning (EFR), which aims to identify these appraisals (or instigators) responsible for an emotion flip. To address EFR, we prepared a new dataset, MELD-I, with the annotations of responsible utterances and respective appraisals. We further compiled the set of instigators in a two-level hierarchy -- coarse-grained with $27$ and fine-grained with $14$ instigators. To benchmark the dataset, we proposed \name\ and performed extensive experiments for the instigator identification in both setups. The comparative analysis against five existing systems showed improvements in the range of $4-11\%$ W-F1 points. Moreover, \name\ adapted well in the fine-grained setup. The EFR task should motivate researchers to ponder more towards the explanability of the emotion dynamics involved in a conversation. The results of the EFR task can be exploited by dialogue agents in order to generate more empathetic response and it can also act as a reward function in the case of reinforcement learning. This is something we would like to explore in the future.

\bibliographystyle{IEEEtran}
\bibliography{affect-ref,anthology,custom}

\begin{thebibliography}{10}
\providecommand{\url}[1]{#1}
\csname url@samestyle\endcsname
\providecommand{\newblock}{\relax}
\providecommand{\bibinfo}[2]{#2}
\providecommand{\BIBentrySTDinterwordspacing}{\spaceskip=0pt\relax}
\providecommand{\BIBentryALTinterwordstretchfactor}{4}
\providecommand{\BIBentryALTinterwordspacing}{\spaceskip=\fontdimen2\font plus
\BIBentryALTinterwordstretchfactor\fontdimen3\font minus
  \fontdimen4\font\relax}
\providecommand{\BIBforeignlanguage}[2]{{%
\expandafter\ifx\csname l@#1\endcsname\relax
\typeout{** WARNING: IEEEtran.bst: No hyphenation pattern has been}%
\typeout{** loaded for the language `#1'. Using the pattern for}%
\typeout{** the default language instead.}%
\else
\language=\csname l@#1\endcsname
\fi
#2}}
\providecommand{\BIBdecl}{\relax}
\BIBdecl

\bibitem{survey}
S.~Poria, N.~Majumder, R.~Mihalcea, and E.~Hovy, ``Emotion recognition in
  conversation: Research challenges, datasets, and recent advances,''
  \emph{IEEE Access}, vol.~7, pp. 100\,943--100\,953, 2019.

\bibitem{social-media1}
K.~Sailunaz and R.~Alhajj, ``Emotion and sentiment analysis from twitter
  text,'' \emph{Journal of Computational Science}, vol.~36, p. 101003, 2019.

\bibitem{social-media2}
L.~Dini and A.~Bittar, ``Emotion analysis on twitter: The hidden challenge,''
  in \emph{Proceedings of the Tenth International Conference on Language
  Resources and Evaluation (LREC'16)}, 2016, pp. 3953--3958.

\bibitem{social-media3}
M.~S. Akhtar, A.~Ekbal, and E.~Cambria, ``How intense are you? predicting
  intensities of emotions and sentiments using stacked ensemble [application
  notes],'' \emph{IEEE Computational Intelligence Magazine}, vol.~15, no.~1,
  pp. 64--75, 2020.

\bibitem{e-commerce}
N.~Gupta, M.~Gilbert, and G.~D. Fabbrizio, ``Emotion detection in email
  customer care,'' \emph{Computational Intelligence}, vol.~29, no.~3, pp.
  489--505, 2013.

\bibitem{erc}
D.~Hazarika, S.~Poria, A.~Zadeh, E.~Cambria, L.-P. Morency, and R.~Zimmermann,
  ``Conversational memory network for emotion recognition in dyadic dialogue
  videos,'' in \emph{Proceedings of the conference. Association for
  Computational Linguistics. North American Chapter. Meeting}, vol. 2018.\hskip
  1em plus 0.5em minus 0.4em\relax NIH Public Access, 2018, p. 2122.

\bibitem{kumar2022discovering}
S.~Kumar, A.~Shrimal, M.~S. Akhtar, and T.~Chakraborty, ``Discovering emotion
  and reasoning its flip in multi-party conversations using masked memory
  network and transformer,'' \emph{Knowledge-Based Systems}, vol. 240, p.
  108112, 2022.

\bibitem{mooren1993contributions}
J.~Mooren and I.~Van~Krogten, ``Contributions to the history of psychology:
  Cxii. magda b. arnold revisited: 1991,'' \emph{Psychological reports},
  vol.~72, no.~1, pp. 67--84, 1993.

\bibitem{lazarus}
R.~S. Lazarus and S.~Folkman, \emph{Stress, appraisal, and coping}.\hskip 1em
  plus 0.5em minus 0.4em\relax Springer publishing company, 1984.

\bibitem{dialog_agent1}
Z.~Lin, A.~Madotto, J.~Shin, P.~Xu, and P.~Fung, ``Moel: Mixture of empathetic
  listeners,'' \emph{arXiv preprint arXiv:1908.07687}, 2019.

\bibitem{dialog_agent2}
J.~Shin, P.~Xu, A.~Madotto, and P.~Fung, ``Generating empathetic responses by
  looking ahead the user’s sentiment,'' in \emph{ICASSP 2020-2020 IEEE
  International Conference on Acoustics, Speech and Signal Processing
  (ICASSP)}.\hskip 1em plus 0.5em minus 0.4em\relax IEEE, 2020, pp. 7989--7993.

\bibitem{dialog_agent3}
Y.~Ma, K.~L. Nguyen, F.~Z. Xing, and E.~Cambria, ``A survey on empathetic
  dialogue systems,'' \emph{Information Fusion}, vol.~64, pp. 50--70, 2020.

\bibitem{ekman}
P.~Ekman, ``An argument for basic emotions,'' \emph{Cognition \& emotion},
  vol.~6, no. 3-4, pp. 169--200, 1992.

\bibitem{picard}
R.~W. Picard, A.~Wexelblat, and C.~I. N.~I. Clifford I.~Nass, ``Future
  interfaces: social and emotional,'' in \emph{CHI'02 Extended Abstracts on
  Human Factors in Computing Systems}, 2002, pp. 698--699.

\bibitem{cowen2017self}
A.~S. Cowen and D.~Keltner, ``Self-report captures 27 distinct categories of
  emotion bridged by continuous gradients,'' \emph{Proceedings of the National
  Academy of Sciences}, vol. 114, no.~38, pp. E7900--E7909, 2017.

\bibitem{mencattini2014speech}
A.~Mencattini, E.~Martinelli, G.~Costantini, M.~Todisco, B.~Basile, M.~Bozzali,
  and C.~Di~Natale, ``Speech emotion recognition using amplitude modulation
  parameters and a combined feature selection procedure,''
  \emph{Knowledge-Based Systems}, vol.~63, pp. 68--81, 2014.

\bibitem{zhang2016intelligent}
L.~Zhang, K.~Mistry, S.~C. Neoh, and C.~P. Lim, ``Intelligent facial emotion
  recognition using moth-firefly optimization,'' \emph{Knowledge-Based
  Systems}, vol. 111, pp. 248--267, 2016.

\bibitem{cui2020eeg}
H.~Cui, A.~Liu, X.~Zhang, X.~Chen, K.~Wang, and X.~Chen, ``Eeg-based emotion
  recognition using an end-to-end regional-asymmetric convolutional neural
  network,'' \emph{Knowledge-Based Systems}, vol. 205, p. 106243, 2020.

\bibitem{liew2016exploring}
J.~S.~Y. Liew and H.~R. Turtle, ``Exploring fine-grained emotion detection in
  tweets,'' in \emph{Proceedings of the NAACL Student Research Workshop}, 2016,
  pp. 73--80.

\bibitem{kaminska2021nearest}
O.~Kaminska, C.~Cornelis, and V.~Hoste, ``Nearest neighbour approaches for
  emotion detection in tweets,'' in \emph{Proceedings of the Eleventh Workshop
  on Computational Approaches to Subjectivity, Sentiment and Social Media
  Analysis}, 2021, pp. 203--212.

\bibitem{SINGH2022245}
Y.~B. Singh and S.~Goel, ``A systematic literature review of speech emotion
  recognition approaches,'' \emph{Neurocomputing}, pp. 245--263, 2022.

\bibitem{THUSEETHAN2022174}
S.~Thuseethan, S.~Rajasegarar, and J.~Yearwood, ``Emosec: Emotion recognition
  from scene context,'' \emph{Neurocomputing}, pp. 174--187, 2022.

\bibitem{9516906}
Y.~Li, T.~Zhang, and C.~L.~P. Chen, ``Enhanced broad siamese network for facial
  emotion recognition in human–robot interaction,'' \emph{IEEE Transactions
  on Artificial Intelligence}, pp. 413--423, 2021.

\bibitem{9736644}
G.~Assunção, B.~Patrão, M.~Castelo-Branco, and P.~Menezes, ``An overview of
  emotion in artificial intelligence,'' \emph{IEEE Transactions on Artificial
  Intelligence}, pp. 867--886, 2022.

\bibitem{icon}
D.~Hazarika, S.~Poria, R.~Mihalcea, E.~Cambria, and R.~Zimmermann, ``Icon:
  interactive conversational memory network for multimodal emotion detection,''
  in \emph{Proceedings of the 2018 conference on empirical methods in natural
  language processing}, 2018, pp. 2594--2604.

\bibitem{erc_knowledge}
P.~Zhong, D.~Wang, and C.~Miao, ``Knowledge-enriched transformer for emotion
  detection in textual conversations,'' in \emph{Proceedings of the 2019
  Conference on Empirical Methods in Natural Language Processing and the 9th
  International Joint Conference on Natural Language Processing
  (EMNLP-IJCNLP)}.\hskip 1em plus 0.5em minus 0.4em\relax Hong Kong, China:
  Association for Computational Linguistics, Nov. 2019, pp. 165--176.

\bibitem{bieru}
W.~Li, W.~Shao, S.~Ji, and E.~Cambria, ``Bieru: bidirectional emotional
  recurrent unit for conversational sentiment analysis,'' \emph{arXiv preprint
  arXiv:2006.00492}, 2020.

\bibitem{dialoguegcn}
D.~Ghosal, N.~Majumder, S.~Poria, N.~Chhaya, and A.~Gelbukh, ``Dialoguegcn: A
  graph convolutional neural network for emotion recognition in conversation,''
  \emph{arXiv preprint arXiv:1908.11540}, 2019.

\bibitem{aghmn}
W.~Jiao, M.~Lyu, and I.~King, ``Real-time emotion recognition via attention
  gated hierarchical memory network,'' in \emph{Proceedings of the AAAI
  Conference on Artificial Intelligence}, vol.~34, no.~05, 2020, pp.
  8002--8009.

\bibitem{tlerc}
D.~Hazarika, S.~Poria, R.~Zimmermann, and R.~Mihalcea, ``Conversational
  transfer learning for emotion recognition,'' \emph{Information Fusion},
  vol.~65, pp. 1--12, 2021.

\bibitem{dialogxl}
W.~Shen, J.~Chen, X.~Quan, and Z.~Xie, ``Dialogxl: All-in-one xlnet for
  multi-party conversation emotion recognition,'' \emph{arXiv preprint
  arXiv:2012.08695}, 2020.

\bibitem{poria2017context}
S.~Poria, E.~Cambria, D.~Hazarika, N.~Majumder, A.~Zadeh, and L.-P. Morency,
  ``Context-dependent sentiment analysis in user-generated videos,'' in
  \emph{ACL}, 2017, pp. 873--883.

\bibitem{jiao2019higru}
W.~Jiao, H.~Yang, I.~King, and M.~R. Lyu, ``Higru: Hierarchical gated recurrent
  units for utterance-level emotion recognition,'' \emph{arXiv preprint
  arXiv:1904.04446}, 2019.

\bibitem{9706271}
G.~Tu, J.~Wen, C.~Liu, D.~Jiang, and E.~Cambria, ``Context- and sentiment-aware
  networks for emotion recognition in conversation,'' \emph{IEEE Transactions
  on Artificial Intelligence}, pp. 699--708, 2022.

\bibitem{yang2022hybrid}
L.~Yang, Y.~Shen, Y.~Mao, and L.~Cai, ``Hybrid curriculum learning for emotion
  recognition in conversation,'' in \emph{Proceedings of the AAAI Conference on
  Artificial Intelligence}, 2022, pp. 11\,595--11\,603.

\bibitem{ma2022multi}
H.~Ma, J.~Wang, H.~Lin, X.~Pan, Y.~Zhang, and Z.~Yang, ``A multi-view network
  for real-time emotion recognition in conversations,'' \emph{Knowledge-Based
  Systems}, p. 107751, 2022.

\bibitem{lian2022smin}
Z.~Lian, B.~Liu, and J.~Tao, ``Smin: Semi-supervised multi-modal interaction
  network for conversational emotion recognition,'' \emph{IEEE Transactions on
  Affective Computing}, 2022.

\bibitem{xlnet}
Z.~Yang, Z.~Dai, Y.~Yang, J.~Carbonell, R.~R. Salakhutdinov, and Q.~V. Le,
  ``Xlnet: Generalized autoregressive pretraining for language understanding,''
  in \emph{Advances in Neural Information Processing Systems}, H.~Wallach,
  H.~Larochelle, A.~Beygelzimer, F.~d\textquotesingle Alch\'{e}-Buc, E.~Fox,
  and R.~Garnett, Eds., vol.~32.\hskip 1em plus 0.5em minus 0.4em\relax Curran
  Associates, Inc., 2019.

\bibitem{LIAN2021483}
Z.~Lian, B.~Liu, and J.~Tao, ``Decn: Dialogical emotion correction network for
  conversational emotion recognition,'' \emph{Neurocomputing}, pp. 483--495,
  2021.

\bibitem{SHOU2022629}
Y.~Shou, T.~Meng, W.~Ai, S.~Yang, and K.~Li, ``Conversational emotion
  recognition studies based on graph convolutional neural networks and a
  dependent syntactic analysis,'' \emph{Neurocomputing}, pp. 629--639, 2022.

\bibitem{emotion-cause1}
S.~Y.~M. Lee, Y.~Chen, and C.-R. Huang, ``A text-driven rule-based system for
  emotion cause detection,'' in \emph{Proceedings of the NAACL HLT 2010
  Workshop on Computational Approaches to Analysis and Generation of Emotion in
  Text}, 2010, pp. 45--53.

\bibitem{emotion-cause2}
L.~Gui, R.~Xu, D.~Wu, Q.~Lu, and Y.~Zhou, ``Event-driven emotion cause
  extraction with corpus construction,'' in \emph{Social Media Content
  Analysis: Natural Language Processing and Beyond}.\hskip 1em plus 0.5em minus
  0.4em\relax World Scientific, 2018, pp. 145--160.

\bibitem{xia2019emotioncause}
R.~Xia and Z.~Ding, ``Emotion-cause pair extraction: A new task to emotion
  analysis in texts,'' 2019.

\bibitem{xia2019rthn}
R.~Xia, M.~Zhang, and Z.~Ding, ``Rthn: A rnn-transformer hierarchical network
  for emotion cause extraction,'' \emph{arXiv preprint arXiv:1906.01236}, 2019.

\bibitem{poria2020recognizing}
S.~Poria, N.~Majumder, D.~Hazarika, D.~Ghosal, R.~Bhardwaj, S.~Y.~B. Jian,
  R.~Ghosh, N.~Chhaya, A.~Gelbukh, and R.~Mihalcea, ``Recognizing emotion cause
  in conversations,'' \emph{arXiv preprint arXiv:2012.11820}, 2020.

\bibitem{meld}
S.~Poria, D.~Hazarika, N.~Majumder, G.~Naik, E.~Cambria, and R.~Mihalcea,
  ``{MELD}: A multimodal multi-party dataset for emotion recognition in
  conversations,'' in \emph{Proceedings of the 57th Annual Meeting of the
  Association for Computational Linguistics}.\hskip 1em plus 0.5em minus
  0.4em\relax Florence, Italy: Association for Computational Linguistics, Jul.
  2019, pp. 527--536.

\bibitem{krippendorff2011computing}
K.~Krippendorff, ``Computing krippendorff's alpha-reliability,'' 2011.

\bibitem{transformer}
\BIBentryALTinterwordspacing
A.~Vaswani, N.~Shazeer, N.~Parmar, J.~Uszkoreit, L.~Jones, A.~N. Gomez, L.~u.
  Kaiser, and I.~Polosukhin, ``Attention is all you need,'' in \emph{Advances
  in Neural Information Processing Systems}, I.~Guyon, U.~V. Luxburg,
  S.~Bengio, H.~Wallach, R.~Fergus, S.~Vishwanathan, and R.~Garnett, Eds.,
  vol.~30.\hskip 1em plus 0.5em minus 0.4em\relax Curran Associates, Inc.,
  2017. [Online]. Available:
  \url{https://proceedings.neurips.cc/paper/2017/file/3f5ee243547dee91fbd053c1c4a845aa-Paper.pdf}
\BIBentrySTDinterwordspacing

\bibitem{chung2014empirical}
J.~Chung, C.~Gulcehre, K.~Cho, and Y.~Bengio, ``Empirical evaluation of gated
  recurrent neural networks on sequence modeling,'' \emph{arXiv preprint
  arXiv:1412.3555}, 2014.

\bibitem{focal:Lin:2017:ICCV}
T.-Y. Lin, P.~Goyal, R.~Girshick, K.~He, and P.~Dollar, ``Focal loss for dense
  object detection,'' in \emph{Proceedings of the IEEE International Conference
  on Computer Vision (ICCV)}, Venice, Italy, Oct 2017.

\bibitem{devlin2018bert}
J.~Devlin, M.-W. Chang, K.~Lee, and K.~Toutanova, ``Bert: Pre-training of deep
  bidirectional transformers for language understanding,'' \emph{arXiv preprint
  arXiv:1810.04805}, 2018.

\bibitem{vaswani2017attention}
A.~Vaswani, N.~Shazeer, N.~Parmar, J.~Uszkoreit, L.~Jones, A.~N. Gomez,
  {\L}.~Kaiser, and I.~Polosukhin, ``Attention is all you need,'' in
  \emph{Advances in neural information processing systems}, 2017, pp.
  5998--6008.

\bibitem{iemocap}
C.~Busso, M.~Bulut, C.-C. Lee, A.~Kazemzadeh, E.~Mower, S.~Kim, J.~N. Chang,
  S.~Lee, and S.~S. Narayanan, ``Iemocap: Interactive emotional dyadic motion
  capture database,'' \emph{Language resources and evaluation}, vol.~42, no.~4,
  pp. 335--359, 2008.

\bibitem{reisenzein1983schachter}
R.~Reisenzein, ``The schachter theory of emotion: two decades later.''
  \emph{Psychological bulletin}, vol.~94, no.~2, p. 239, 1983.

\end{thebibliography}

\begin{IEEEbiography}[{\includegraphics[width=1in,height=1.25in,clip,keepaspectratio]{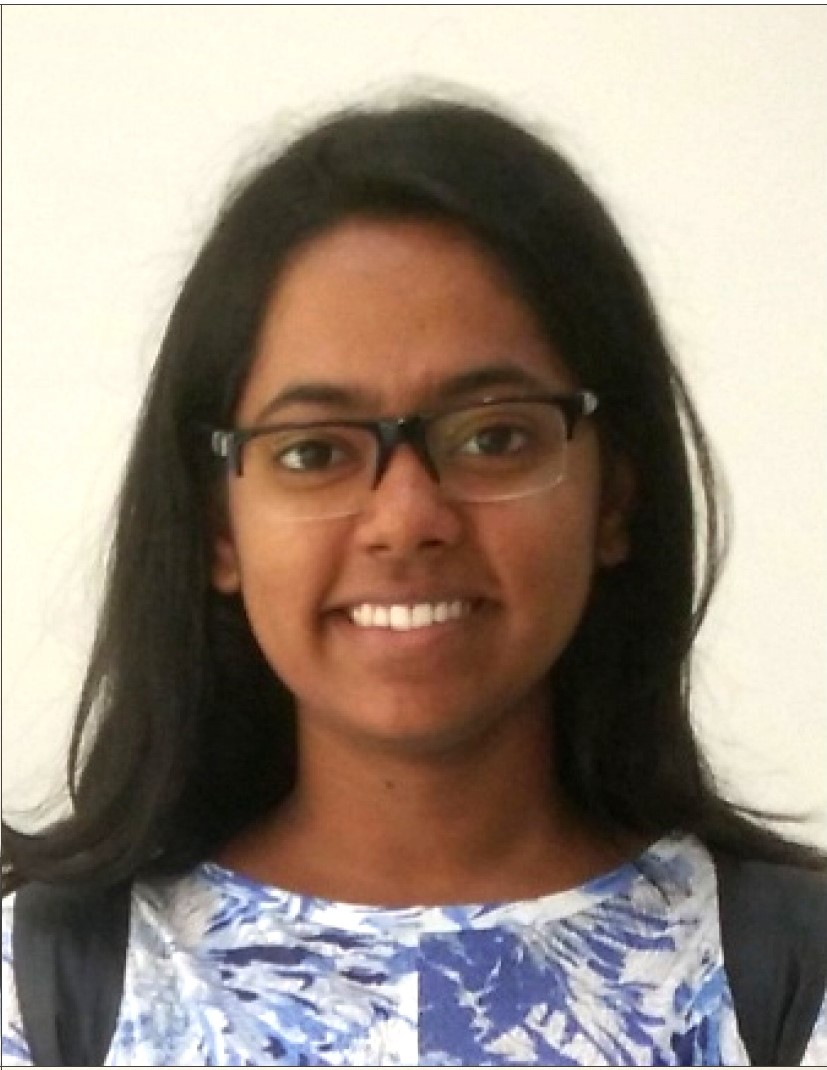}}]{Shivani Kumar}{\space} is a PhD scholar at Indraprastha Institute of Information Technology Delhi (IIIT Delhi), India. She holds a Senior Research Fellowship and works in the domain of Natural Language Processing, primarily in the area of understanding and explaining various affects, like emotions, sarcasm, and humour, in conversational data. 
\end{IEEEbiography}

\begin{IEEEbiography}[{\includegraphics[width=1in,height=1.25in,clip,keepaspectratio]{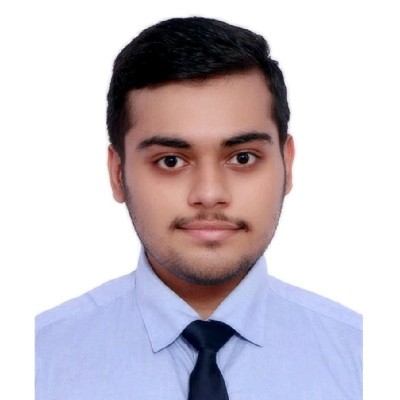}}]{Shubham Dudeja}{\space} was a masters student at Indraprastha Institute of Information Technology Delhi (IIIT Delhi), India for the duration of this work. His research interest lies in Natural Language Processing and its sub-fields. 
\end{IEEEbiography}

\begin{IEEEbiography}[{\includegraphics[width=1in,height=1.25in,clip,keepaspectratio]{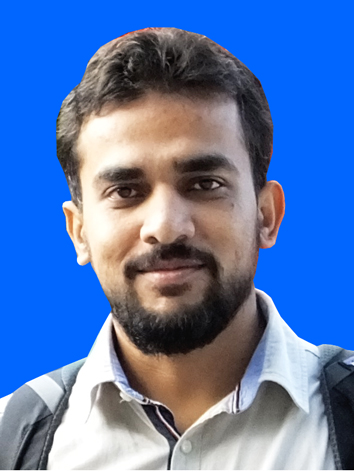}}]{Md Shad Akhtar}{\space} is currently an Assistant Professor at Indraprastha Institute of Information Technology Delhi (IIIT Delhi). His main area of research is NLP with a focus on the affective analysis. He completed his PhD form IIT Patna.
\end{IEEEbiography}

\begin{IEEEbiography}[{\includegraphics[width=1in,height=1.25in,clip,keepaspectratio]{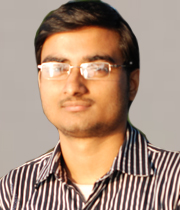}}]{{Tanmoy Chakraborty}}{\space} is an Associate Professor in the Dept. of EE, Indian Institute of Technology Delhi, India since September 2022. Before joining IIT Delhi, he served as an Associate Professor in the Dept of CSE, IIIT Delhi, India. He completed his postdoctoral research from University of Maryland, College Park after obtaining his PhD from the Dept. of CSE, IIT Kharagpur, India as a Google PhD scholar. His broad research interests include NLP, Graph Neural Networks, and Social Computing.
\end{IEEEbiography}


\section{Supplementary}
\subsection{Background of Instigator Labels}
\label{sec:app_inst_bg}
We studied conventional emotion theories to curate a specially designed instigator set for all types of emotion flips. Magda Arnold \cite{mooren1993contributions} proposed a theory of emotion where she coined the word `appraisal', which refers to the cognitive processes prior to emotion evocation. The ``cognitive theory" \cite{mooren1993contributions} claims that before a person can experience an emotion, he/she must make an appraisal of the situation. According to Arnold, this initial appraisal results in an emotion which is exhibited via both physiological and psychological experiences. Psychologist Richard Lazarus extended Arnold's research and developed his cognitive-mediational theory \cite{lazarus}. It adds that emotions are a result of appraisals to stimuli, following which emotions are induced in a person. He argued that the emotional and physiological arousals happen at the same time and not one after the other as was suggested in Schachter-Singer theory \cite{reisenzein1983schachter}. Lazarus further stressed that the intensity of emotions is commanded by psychological processes that decide how a person will react to his/her environment, thus becoming a basis of their emotional reaction. In his theory, Lazarus proposed two types of appraisals -- primary and secondary. The primary appraisal is responsible for associating a meaning behind an event, while the secondary appraisal assesses an individual's ability to cope up with the consequences of the event. His theory was primarily based on the context of stress. Consequently, the primary appraisals could be seen as the judgment induced to judge the extent of potential harm that a (stressful) event might introduce. This judgment then sets off the secondary appraisal, where the individual finds ways to deal with the (stressful) event.

Furthermore, for the domain of stressful scenarios, Lazarus defined three primary appraisals -- threat, challenge, and harm/loss. We aim to extend this set of appraisals in a more general fashion in our work. That is, we aim to come up with a set of appraisals or instigators to reason any emotion change (e.g., \textit{joy} $\rightarrow$ \textit{anger}). Since we consider Ekman's set of emotions \cite{ekman} in our study, we want a set of instigators that are able to quantify all possible $42$ flips of emotion ($7 \times 6$ emotion flip pairs). 

Following the guidelines mentioned in Section III of the main text, we come up with a set of $27$ possible instigators which are able to reason all the flips possible in Ekman's emotion set shown in Table \ref{tab:instigators}.
Division of the instigator labels into the set of positive, negative, and ambiguous can be seen in Table \ref{tab:inst_div}. Further, 
Table \ref{tab:inst_dist_emo} present the set of possible instigators for each emotion pairs (rows: source, column: target).

\begin{table}[t]
\centering
\caption{Instigator labels with their definitions.}
\resizebox{\columnwidth}{!}{%
\begin{tabular}{|l|l|l|}
\hline
\multicolumn{2}{|c|}{\textbf{Instigator}} & \\ \cline{1-2}
\multicolumn{1}{|c|}{\textbf{Coarse-grained}} & \multicolumn{1}{c|}{\textbf{Fine-grained}} & \multicolumn{1}{c|}{\multirow{-2}{*}{\textbf{Definition}}} \\ \hline

\hline
\multirow{2}{*}{Annoyance} & Annoyance & To get irritated by something/someone \\ \cline{2-3} 
 & Pain & An unpleasant physical sensation \\ \hline
Awkwardness & Awkwardness & Feeling embarrassed \\ \hline
Benefit & Benefit & Anything profitable/advantageous \\ \hline
\multirow{2}{*}{Cheer} & Cheer & Being encouraged/comforted \\ \cline{2-3} 
 & Humor & Being amused \\ \hline
Confusion & Confusion & Feeling uncertain about the situation \\ \hline
Curiosity & Curiosity & Feeling inquisitive \\ \hline
\multirow{2}{*}{Ease} & Calmness & Not feeling agitated/ Feeling relaxed \\ \cline{2-3} 
 & Relief & Feeling rid of anxiety \\ \hline
\multirow{3}{*}{Excitement} & Excitement & Being eager or enthusiastic \\ \cline{2-3} 
 & Satisfaction & Fulfilment of wishes/expectations \\ \cline{2-3} 
 & Desire & Wanting something to happen \\ \hline
\multirow{2}{*}{Dazzle} & Adoration & Feeling of deep love and respect \\ \cline{2-3} 
 & Impressed & Admiration for something/someone \\ \hline
Loss & Loss & Suffer. Losing someone/something \\ \hline
\multirow{3}{*}{Nervousness} & Nervousness & Feeling anxious \\ \cline{2-3} 
 & Scold & Be criticised angrily \\ \cline{2-3} 
 & Guilt & Feeling morally wrong \\ \hline
Shock & Shock & Sudden upsetting/surprising event \\ \hline
\multirow{3}{*}{Threat} & Threat & Statement to inflict pain/damage \\ \cline{2-3} 
 & Horror & Feeling of extreme terror \\ \cline{2-3} 
 & Abuse & Usage of abusive language \\ \hline
\multirow{4}{*}{Others} & Boredom & Feeling uninterested \\ \cline{2-3} 
 & Sympathy & Feeling pity for someone’s misfortune \\ \cline{2-3} 
 & Challenge & Contest/justify something \\ \cline{2-3} 
 & Nostalgia & Remembering the past \\ \hline
\end{tabular}%
}
\label{tab:instigators}
\vspace{-3mm}
\end{table}
\begin{table}[t]
\centering
\caption{Division of 27 instigators into positive, negative, and ambiguous instigators.}
\resizebox{\columnwidth}{!}{%
\begin{tabular}{c|c|p{20em}|}
\cline{2-3}
& & \multicolumn{1}{c|}{\bf Instigator} \\ \cline{2-3} 
\multirow{6}{*}{\rotatebox{90}{\textbf{Sentiment}}} & \multicolumn{1}{c|}{\textbf{Positive}} & adoration, benefit, calmness, cheer, desire, excitement, humour, impressed, relief, satisfaction \\ \cline{2-3} 
& \multicolumn{1}{c|}{\textbf{Negative}} & abuse, annoyance, guilt, horror, loss, nervousness, pain, scold, shock, sympathy, threat \\ \cline{2-3} 
& \multicolumn{1}{c|}{\textbf{Ambiguous}} & awkwardness, boredom, challenge, confusion, curiosity, nostalgia \\ \cline{2-3}
\end{tabular}%
}
\label{tab:inst_div}
\vspace{-5mm}
\end{table}
{\tiny
\begin{table*}[ht]
\centering
\caption{Possible instigators with respect to the emotion flips. This trend is observed from the ground-truth EFR labels in MELD-I. A cell $(i,j)$  represents the possible instigators responsible for the emotion flip from $i$ to $j$. E.g., \textit{horror}, \textit{nervousness}, and \textit{threat} are the possible instigators that can result in the emotion flip from \textit{disgust} to \textit{fear}.}
\resizebox{\textwidth}{!}
{%
\begin{tabular}{c|r|p{20em}|p{12em}|p{20em}|p{20em}|}
 \multicolumn{1}{c}{} & \multicolumn{1}{c}{} & \multicolumn{4}{c}{\textbf{Target}} \\ \cline{3-6}
 \multicolumn{1}{c}{} & & \multicolumn{1}{c|}{\textit{neutral}} & \multicolumn{1}{c|}{\textit{disgust}} & \multicolumn{1}{c|}{\textit{sadness}} & \multicolumn{1}{c|}{\textit{joy}} \\ \cline{2-6} 

\multirow{27}{*}{\rotatebox{90}{\bf Source}} & \multirow{4}{*}{\textit{{neutral}}} & & annoyance, awkwardness, boredom, calmness, challenge, confusion, curiosity, desire, horror, loss, shock, threat & annoyance, awkwardness, calmness, challenge, confusion, curiosity, desire, excitement, guilt, horror, loss, nervousness, nostalgia, shock, sympathy, threat & adoration, annoyance, awkwardness, benefit, calmness, challenge, cheer, confusion, curiosity, desire, excitement, humour, impressed, nervousness, nostalgia, relief, satisfaction, shock, sympathy \\ \cline{2-6} 
 & \multirow{3}{*}{\textit{{disgust}}} & annoyance, awkwardness, benefit, calmness, challenge, cheer, confusion, curiosity, desire, excitement, guilt, nervousness, scold, sympathy & & annoyance, awkwardness, confusion, loss, sympathy, threat & adoration, awkwardness, benefit, cheer, confusion, curiosity, excitement, horror, impressed, relief \\ \cline{2-6} 
 & \multirow{4}{*}{\textit{{sadness}}} & adoration, annoyance, awkwardness, boredom, calmness, challenge, cheer, confusion, curiosity, excitement, guilt, impressed, loss, nervousness, nostalgia, relief, satisfaction, sympathy & abuse, annoyance, awkwardness, confusion, loss & & adoration, annoyance, awkwardness, benefit, calmness, challenge, cheer, confusion, curiosity, desire, excitement, humour, impressed, relief, satisfaction \\ \cline{2-6} 
 & \multirow{5}{*}{\textit{{joy}}} & adoration, annoyance, awkwardness, benefit, boredom, calmness, challenge, cheer, confusion, curiosity, desire, excitement, guilt, humour, impressed, loss, nervousness, nostalgia, relief, satisfaction, sympathy & annoyance, awkwardness, calmness, challenge, curiosity, excitement, humour, impressed, loss, nervousness, threat & adoration, annoyance, awkwardness, calmness, cheer, confusion, curiosity, desire, excitement, guilt, loss, nervousness, sympathy, threat &  \\ \cline{2-6} 
 & \multirow{4}{*}{\textit{{surprise}}} & annoyance, awkwardness, calmness, challenge, cheer, confusion, curiosity, desire, excitement, guilt, humour, impressed, loss, nervousness, relief, satisfaction, shock, sympathy, threat & annoyance, awkwardness, challenge, confusion, horror, loss, nervousness, nostalgia, shock, sympathy, threat & adoration, annoyance, awkwardness, benefit, calmness, cheer, confusion, curiosity, excitement, guilt, horror, loss, nervousness, nostalgia, scold, shock, sympathy, threat & adoration, annoyance, awkwardness, benefit, calmness, challenge, cheer, confusion, curiosity, desire, excitement, humour, impressed, nervousness, nostalgia, relief, satisfaction, shock \\ \cline{2-6} 
 & \multirow{3}{*}{\textit{{fear}}} & adoration, annoyance, awkwardness, calmness, cheer, confusion, curiosity, excitement, nervousness, relief, satisfaction, sympathy, threat & awkwardness & annoyance, awkwardness, calmness, challenge, confusion, desire, guilt, horror, humour, loss, nervousness, shock, sympathy, threat & adoration, benefit, cheer, curiosity, desire, excitement, humour, nervousness, sympathy \\ \cline{2-6} 
 & \multirow{4}{*}{\textit{{anger}}} & annoyance, awkwardness, benefit, boredom, calmness, challenge, cheer, confusion, curiosity, desire, excitement, guilt, impressed, loss, nervousness, relief, satisfaction, sympathy, threat & annoyance, awkwardness, calmness, challenge, cheer, loss, sympathy, threat & adoration, annoyance, awkwardness, calmness, challenge, cheer, confusion, curiosity, excitement, guilt, horror, impressed, loss, nervousness, scold, shock, sympathy, threat & adoration, annoyance, awkwardness, benefit, calmness, challenge, cheer, confusion, curiosity, desire, excitement, guilt, humour, impressed, satisfaction \\ \cline{2-6} 
 \multicolumn{6}{c}{} \\
\end{tabular}%
}
\resizebox{0.9\textwidth}{!}
{%
\begin{tabular}{crlll}
 & \multicolumn{1}{c}{\textit{}} & \multicolumn{3}{c}{\textbf{Target}} \\ \cline{3-5}
\textit{} &  & \multicolumn{1}{|c|}{\textit{surprise}} & \multicolumn{1}{c|}{\textit{fear}} & \multicolumn{1}{c|}{\textit{anger}} \\ \cline{2-5} 
\multirow{23}{*}{\rotatebox{90}{\bf Source}} & \multicolumn{1}{|r|}{\textit{neutral}} & \multicolumn{1}{p{20em}|}{adoration, annoyance, awkwardness, benefit, boredom, calmness, challenge, cheer, confusion, curiosity, desire, excitement, horror, humour, impressed, loss, nervousness, relief, shock, threat} & \multicolumn{1}{p{20em}|}{annoyance, awkwardness, benefit, calmness, challenge, confusion, curiosity, desire, excitement, guilt, horror, nervousness, pain, scold, shock, threat} & \multicolumn{1}{p{20em}|}{abuse, annoyance, awkwardness, boredom, calmness, challenge, cheer, confusion, curiosity, desire, excitement, guilt, horror, humour, loss, nervousness, relief, scold, shock, sympathy, threat} \\ \cline{2-5} 
 & \multicolumn{1}{|r|}{\textit{disgust}} & \multicolumn{1}{p{20em}|}{annoyance, awkwardness, boredom, calmness, challenge, confusion, curiosity, excitement, shock} & \multicolumn{1}{p{20em}|}{horror, nervousness, threat} & \multicolumn{1}{p{20em}|}{annoyance, awkwardness, challenge, confusion, loss, scold, sympathy, threat} \\ \cline{2-5} 
 & \multicolumn{1}{|r|}{{\textit{sadness}}} & \multicolumn{1}{p{20em}|}{annoyance, awkwardness, calmness, challenge, cheer, confusion, curiosity, desire, excitement, guilt, horror, impressed, loss, shock, sympathy, threat} & \multicolumn{1}{p{20em}|}{annoyance, awkwardness, calmness, challenge, confusion, guilt, horror, loss, nervousness, shock, sympathy, threat} & \multicolumn{1}{p{20em}|}{annoyance, awkwardness, calmness, challenge, confusion, curiosity, desire, excitement, guilt, horror, loss, nervousness, shock, sympathy, threat} \\ \cline{2-5} 
 & \multicolumn{1}{|r|}{{\textit{joy}}} & \multicolumn{1}{p{20em}|}{adoration, annoyance, awkwardness, benefit, calmness, challenge, cheer, confusion, curiosity, excitement, guilt, horror, humour, impressed, nervousness, satisfaction, shock, threat} & \multicolumn{1}{p{20em}|}{awkwardness, calmness, cheer, desire, excitement, horror, loss, nervousness, threat} & \multicolumn{1}{p{20em}|}{abuse, annoyance, awkwardness, benefit, calmness, challenge, cheer, confusion, curiosity, excitement, guilt, horror, loss, nervousness, scold, shock, sympathy, threat} \\ \cline{2-5} 
 & \multicolumn{1}{|r|}{{\textit{surprise}}} & \multicolumn{1}{p{20em}|}{} & \multicolumn{1}{p{20em}|}{awkwardness, calmness, challenge, desire, excitement, guilt, horror, nervousness, threat} & \multicolumn{1}{p{20em}|}{annoyance, awkwardness, calmness, cheer, confusion, curiosity, excitement, guilt, horror, loss, nervousness, relief, scold, shock, sympathy, threat} \\ \cline{2-5} 
 & \multicolumn{1}{|r|}{{\textit{fear}}} & \multicolumn{1}{p{20em}|}{annoyance, awkwardness, benefit, calmness, cheer, curiosity, excitement, guilt, humour, nervousness, shock, threat} & \multicolumn{1}{p{20em}|}{} & \multicolumn{1}{p{20em}|}{annoyance, awkwardness, calmness, confusion, curiosity, excitement, guilt, horror, loss, nervousness, scold, threat} \\ \cline{2-5} 
 & \multicolumn{1}{|r|}{{\textit{anger}}}  & \multicolumn{1}{p{20em}|}{annoyance, awkwardness, calmness, confusion, curiosity, excitement, guilt, horror, loss, nervousness, pain, relief, scold, shock, sympathy, threat} & \multicolumn{1}{p{20em}|}{annoyance, awkwardness, challenge, guilt, horror, loss, nervousness, threat} & \multicolumn{1}{p{20em}|}{} \\ \cline{2-5} 
\end{tabular}%
}
\label{tab:inst_dist_emo}
\end{table*}
}

\end{document}